\documentclass[10pt,twocolumn,letterpaper]{article}

\usepackage[pagenumbers]{paperstyle}

\newcommand{\red}[1]{{\color{red}#1}}
\newcommand{\todo}[1]{{\color{red}#1}}
\newcommand{\TODO}[1]{\textbf{\color{red}[TODO: #1]}}

\usepackage{colortbl}
\definecolor{resultgroupgray}{HTML}{E4E4E4}
\definecolor{resultuppergray}{HTML}{FFFFFF}
\definecolor{resultattention}{HTML}{FFFFFF}
\definecolor{resultmerge}{HTML}{FFFFFF}
\definecolor{resultadaptive}{HTML}{FFFFFF}
\definecolor{resultours}{HTML}{F1ECF7}
\definecolor{resultreduction}{HTML}{74569B}

\makeatletter
\IfFormatAtLeastTF{2025-06-01}{%
  \@ifpackageloaded{lineno}{%
    \RemoveFromHook{build/column/before}[lineno]%
    \AddToHook{build/column/before}[paper-lineno-column-fix]{%
      \if@twocolumn
        \if@firstcolumn
          \protected@write\@auxout{}{\string\def\string\@LN@column{2}}%
        \else
          \protected@write\@auxout{}{\string\def\string\@LN@column{1}}%
        \fi
      \fi
    }%
  }{}%
}{}%
\makeatother

\definecolor{resultours}{HTML}{FFF1E6}
\definecolor{resultreduction}{HTML}{D9772E}
\usepackage{float}
\usepackage{placeins}
\usepackage{algorithm}
\usepackage[noend]{algpseudocode}
\usepackage{cuted}
\usepackage{wrapfig}

\definecolor{linkblue}{rgb}{0.21,0.49,0.74}
\usepackage[breaklinks,colorlinks,allcolors=linkblue]{hyperref}
\makeatletter
\def\theHALG@line{\thealgorithm.\arabic{ALG@line}}
\makeatother

\algrenewcommand\algorithmicrequire{\textbf{Input:}}
\algrenewcommand\algorithmicensure{\textbf{Output:}}

\newcommand{\methodname}{S\textsuperscript{2}Prune\xspace}
\newcommand{\datasetSQA}{SQA\textsuperscript{IMG}}
\newcommand{\datasetVQAText}{VQA\textsuperscript{Text}}
\newcommand{\datasetMMBEN}{MMB\textsuperscript{EN}}
\newcommand{\datasetMMBCN}{MMB\textsuperscript{CN}}

\title{S\textsuperscript{2}Prune: Spatially Structured Visual Token Pruning for Multimodal Large Language Models}

\author{Yuanyuan Jia \qquad Shunpu Tang \qquad Qianqian Yang\\
College of Information Science and Electronic Engineering, Zhejiang University\\
Hangzhou, China\\
{\tt\small \{yuanyuanjia,tangshunpu,qianqianyang20\}@zju.edu.cn}
}

\begin{document}
\maketitle
\raggedbottom
\begin{abstract}
Visual token pruning reduces the inference overhead of multimodal large language
models (MLLMs) by retaining only a subset of visual tokens. Existing methods
usually select tokens based on importance or redundancy. However, we observe
that these criteria produce stable spatial biases across inputs and do not always
outperform simple Uniform Grid sampling, highlighting the value of broad spatial
coverage. Motivated by this, we propose \textbf{\methodname}, a training-free pruning
method that preserves spatial coverage while adapting token density to local
image structure. We first divide the image into regions and assign at least one
token to each region to preserve coverage. The remaining token budget is then
distributed according to Laplacian variation, giving more tokens
to regions with richer structure. We then use Early Representation Change (ERC), computed
from the first decoder block, to select representative tokens within each
region. We evaluate \methodname across diverse settings and two MLLM architectures. On
Qwen2.5-VL-7B-Instruct, it achieves the highest average accuracy among the
evaluated training-free pruning methods. With only 32 of the original 576 visual tokens, it still retains 79.3\% of the full-model performance.
Code is available at
\href{https://github.com/yuanyuanjia71-spec/S2Prune.git}{github.com/yuanyuanjia71-spec/S2Prune}.
\end{abstract}

\section{Introduction}
\label{sec:introduction}

\begin{figure}[t]
  \centering
  \includegraphics[width=\columnwidth]{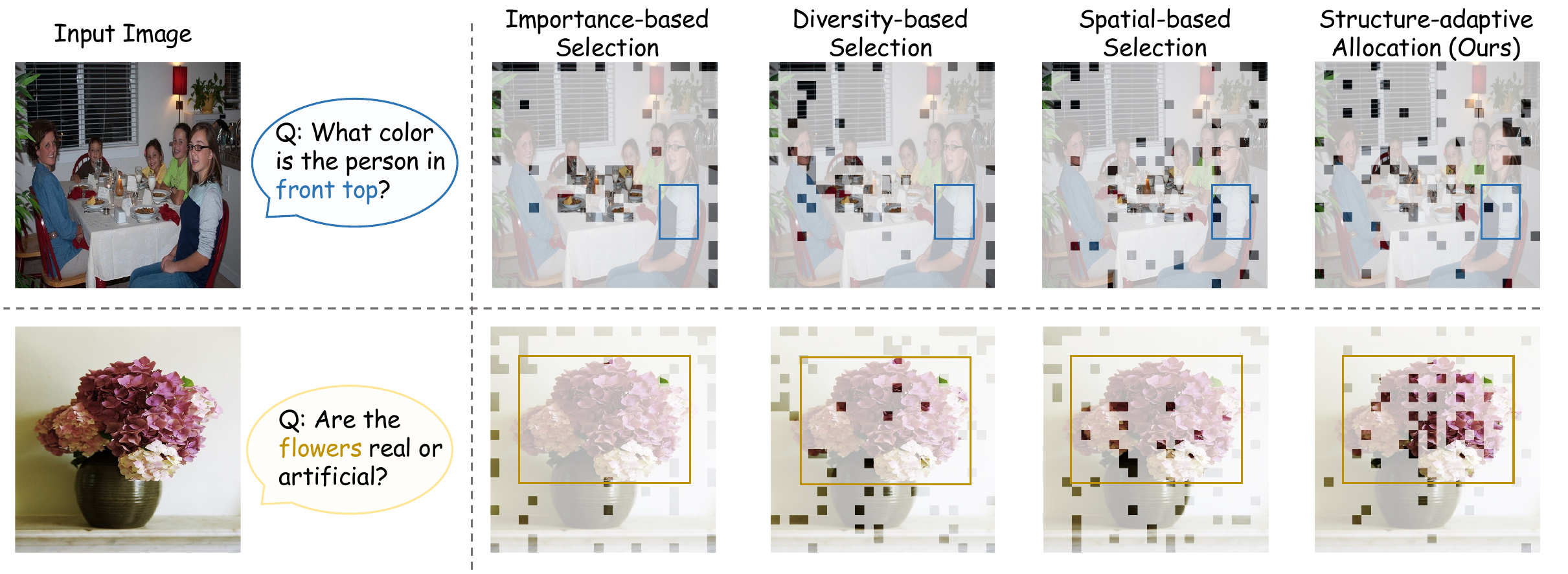}
  \caption{Comparison of four visual token selection strategies. Attention-
and diversity-based methods exhibit spatial bias, while spatial-based
selection preserves coverage but can miss important local details.
\textbf{\methodname} preserves broad coverage and allocates more tokens to
structurally rich regions.}
\label{fig:spatial-capacity-motivation}
\end{figure}

Recently, multimodal large language models (MLLMs) have achieved strong performance in visual understanding and multimodal reasoning. Most MLLMs encode visual inputs into sequences of visual tokens and process them together with text tokens using a large language model. To preserve fine-grained visual information, especially for high-resolution inputs, a large number of visual tokens are often required. These long visual sequences significantly increase Transformer prefill computation, KV-cache usage, and inference latency. To address these issues, visual token pruning has been widely studied to keep only a small subset of visual tokens while preserving model performance~\cite{chen2024fastv,endo2025feather,yang2025visionzip}.

Early visual token pruning methods mainly focus on token importance or
diversity. Importance-based methods evaluate how useful each token is, for
example using attention scores or text relevance, and retain tokens with high
scores~\cite{chen2024fastv}. Diversity-based methods instead compare visual
tokens with one another and remove tokens that contain repeated or similar
information~\cite{wen2025dart,jiang2025gprune}. These two approaches are
complementary and can also be combined. Recent studies further incorporate
spatial information into visual token pruning, for example by encouraging
retained tokens to cover different image locations, assigning different token
budgets to different regions, or correcting the model's preference for specific
spatial positions
~\cite{yang2025topv,ye2025atpllava,zou2025holov,
duan2025gridprune,wang2026posprune,zhang2026d2pruner}.

Despite these advances, recent studies show that elaborate token-selection criteria do not always improve downstream performance~\cite{wen2025rightproblem,wang2026random,endo2025feather}, while simple uniform sampling can improve pruning by maintaining broader spatial coverage. This raises a natural question: \emph{why can such a simple spatial pattern work so well?}

\begin{figure*}[t]
  \centering
  \begin{subfigure}[t]{0.365\textwidth}
    \centering
    \includegraphics[width=\linewidth]{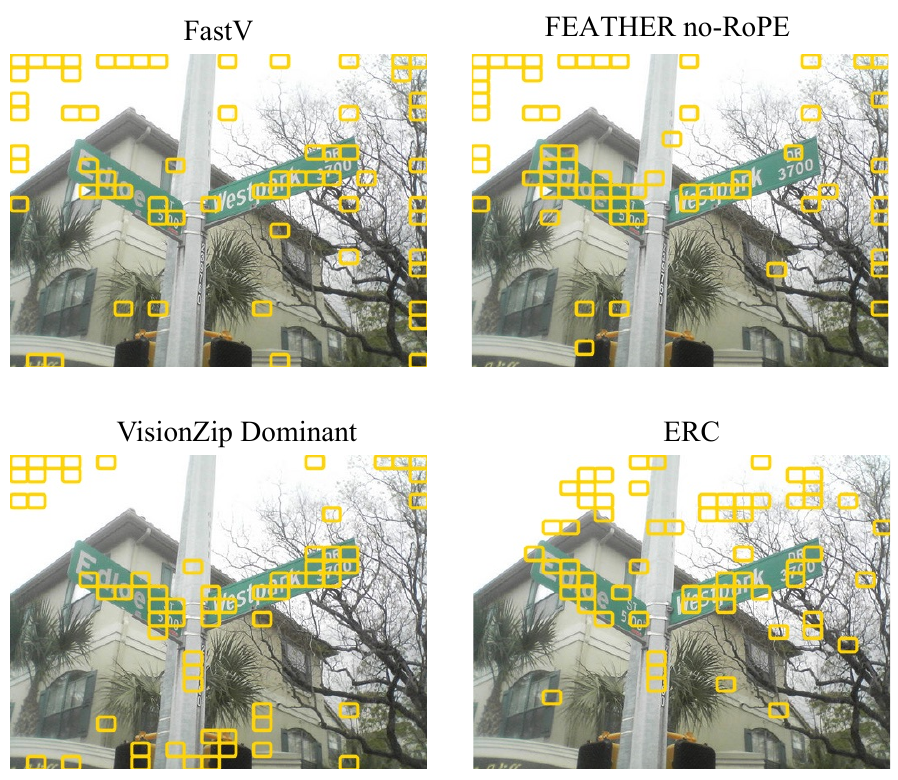}
    \caption{Token selections on the same VQAv2 example~\cite{goyal2017vqa}.}
    \label{fig:importance-allocation-qualitative}
  \end{subfigure}\hfill
  \begin{subfigure}[t]{0.333\textwidth}
    \centering
    \includegraphics[width=\linewidth]{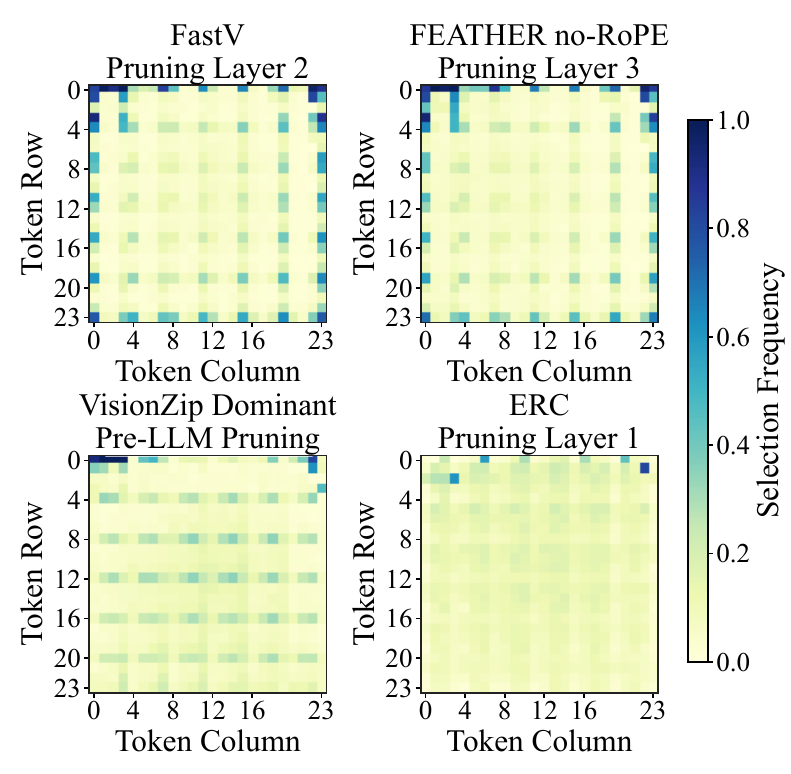}
    \caption{Token selection frequencies on \datasetMMBEN.}
    \label{fig:importance-allocation-heatmaps}
  \end{subfigure}\hfill
  \begin{subfigure}[t]{0.28\textwidth}
    \centering
    \includegraphics[width=\linewidth]{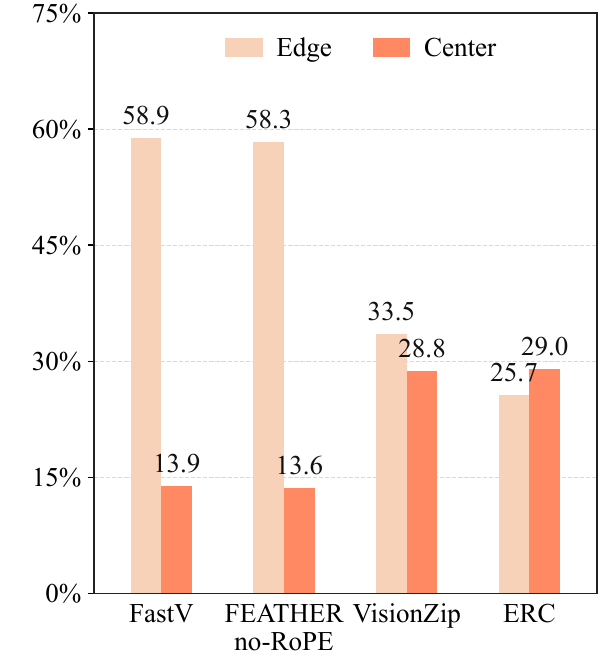}
    \caption{Edge and center allocation ratios.}
    \label{fig:importance-allocation-ratios}
  \end{subfigure}
  \caption{Spatial allocations induced by different token scores under Global
  Top-$B$ selection ($B=64$). (a) Selections for the same VQAv2 example.
  (b) Selection frequencies over 1,000 \datasetMMBEN{}
  samples~\cite{liu2023mmbench}. (c) Retained-token fractions in the edge and
  center regions.}
  \label{fig:importance-spatial-allocations}
\end{figure*}

Our preliminary analysis helps explain this. We fix the model, input, token
budget, and Top-$B$ selection rule, and change only the token score. Different
scoring criteria produce distinct spatial biases, and these biases remain
stable across many image--question pairs. Thus, global token scoring determines
not only \emph{which} tokens are retained, but also \emph{where} the limited
token budget is spent. We further compare these score-driven patterns with
Uniform Grid sampling. By spreading tokens across the image, Uniform Grid
preserves broad spatial support and remains competitive under aggressive
pruning. However, it uses the same token density everywhere and ignores local
image structure. Broad coverage does not require every image region to receive
the same number of tokens: smooth regions can often be represented sparsely,
whereas text, object boundaries, thin structures, and fine textures may require
denser sampling.

This leads to an allocation problem similar to image compression. Under a fixed bitrate, smooth regions require fewer bits than regions with richer local structure. Likewise, under a fixed visual-token budget, different image regions may require different numbers of tokens to preserve their content. Therefore, a good pruning strategy should preserve broad spatial coverage while adapting the token density to local image structure, as illustrated in Fig.~\ref{fig:spatial-capacity-motivation}.

Motivated by this, we formulate visual token pruning as spatial rate allocation
in the token domain and introduce \methodname, a training-free pruning
framework. Given a fixed budget, \methodname first divides the image into
regions and assigns at least one token to each region to preserve coverage. We
use the variance of Laplacian responses to measure regional structural
complexity and allocate the remaining budget accordingly, giving more tokens
to regions with richer structure. Once the regional budgets are fixed, we use
\emph{Early Representation Change} (ERC), computed from the first decoder
block, to select representative tokens within each region. Thus, regional
structure determines \emph{how many} tokens each region receives, while ERC
determines \emph{which} tokens are selected.

We evaluate \methodname on ten multimodal benchmarks, multiple token budgets, and different MLLM architectures. On Qwen2.5-VL-7B-Instruct, \methodname achieves the highest average accuracy among the evaluated training-free pruning methods at budgets of 128, 64, and 32 tokens. With only 32 of the original 576 visual tokens, it reaches 55.9 average accuracy, outperforming the strongest baseline by 0.9 points while retaining $79.3\%$ of the full-model performance.
\begin{itemize}
    \item We revisit visual token pruning from a spatial allocation perspective.
    Our analysis shows that global token scoring can induce very different
    spatial allocations, motivating a simple principle: preserve broad spatial
    coverage while adapting token density to local image structure.

    \item We propose \methodname, a training-free pruning framework that follows
    this principle. Laplacian variation determines how many tokens each region
    receives, while Early Representation Change (ERC) uses the first decoder
    response to select representative tokens within each region.

    \item We evaluate \methodname on ten multimodal benchmarks, multiple token
    budgets, and different MLLM architectures. The results demonstrate
    consistent improvements over existing training-free pruning methods,
    especially under aggressive token pruning.
\end{itemize}

\section{Related Work}
\label{sec:related-work}

\subsection{Vision Language Models}
\label{sec:related-work-lvlms}

\begingroup
\hyphenpenalty=10000
\exhyphenpenalty=10000
Vision language models (VLMs) connect pretrained language models to
visual encoders through an alignment module, enabling a single model to process
images and text~\cite{alayrac2022flamingo,li2023blip2,liu2023llava,bai2023qwenvl}.
Representative models include LLaVA, Qwen-VL, and
InternVL~\cite{liu2023llava,bai2023qwenvl,chen2024internvl}. Recent models such
as LLaVA-NeXT and Qwen2.5-VL support high-resolution inputs for fine-grained
visual tasks~\cite{liu2024llavanext,bai2025qwen25vl}, but produce longer visual
token sequences. These tokens increase prefill latency and KV-cache
memory~\cite{yang2025visionzip}, while attention cost grows quadratically with
sequence length~\cite{vaswani2017attention}. Visual token pruning reduces these
costs by removing redundant tokens while preserving task-relevant
information.
\par
\endgroup

\subsection{Token Pruning for VLMs}
\label{sec:related-work-token-pruning}

Visual token pruning reduces VLM inference cost by removing redundant visual
representations without modifying the underlying architecture. Earlier
vision-Transformer work explored dynamic token sparsification, token
reorganization, and similarity-based
merging~\cite{rao2021dynamicvit,liang2022evit}.

For VLMs, attention-based methods retain tokens using decoder attention or
cross-modal relevance~\cite{chen2024fastv,zhang2025sparsevlm,huang2024ivtp,
sun2025lvpruning,endo2025feather}. Redundancy- and relation-based methods prune,
merge, or rank visual features according to duplication or feature
relations~\cite{wen2025dart,jiang2025gprune,shang2025prumerge,dhouib2025pact}.
Other approaches construct a more representative subset through diversity, or
combine visual importance with similarity- or diversity-based
complementarity~\cite{alvar2025divprune,yang2025visionzip,
zhang2025vispruner,zhang2025cdpruner,li2025mob}.

Recent studies increasingly consider the spatial organization of retained
visual tokens. FEATHER~\cite{endo2025feather} improves spatial coverage through
uniform sampling, TopV~\cite{yang2025topv} incorporates spatial distance, and
ATP-LLaVA~\cite{ye2025atpllava} adapts pruning across layers and inputs.
GridPrune~\cite{duan2025gridprune} further separates query-conditioned regional
budget allocation from local token selection. These works demonstrate that
spatial organization is an important factor beyond token-wise importance.

Our work considers a different role of spatial allocation. Rather than using
regional budgets to indicate \emph{where task-relevant evidence lies}, we use
them to model \emph{how much representation capacity a region requires}.
Semantic relevance and representation demand are not necessarily aligned:
regions of comparable relevance may require substantially different token
densities to preserve their visual structure. \methodname therefore allocates
capacity according to image-intrinsic structural variation and uses
decoder-derived ERC only for local representative selection.


%

\begin{figure}[t]
  \centering
  \includegraphics[width=0.88\columnwidth]{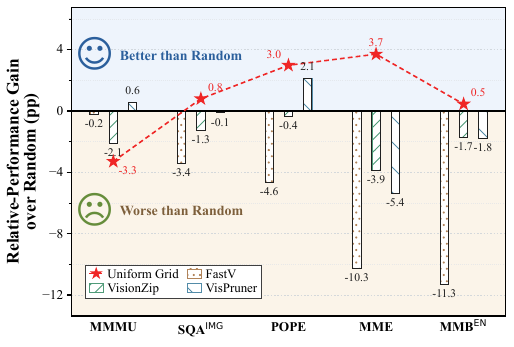}
  \caption{Performance-retention differences from Random on Qwen2.5-VL-7B at
  $B=64$. Values are percentage-point differences in retention relative to the
  unpruned 576-token model; zero denotes Random. Uniform Grid improves four of
  five benchmarks.}
  \label{fig:uniform-vs-baselines}
\end{figure}

\section{Preliminary and Motivation}
\label{sec:background}

\subsection{Preliminary}

\noindent\textbf{Architecture of VLMs.} A typical VLM consists of a vision
encoder, a vision-language projector, and a large language model
(LLM)~\cite{li2023blip2,liu2023llava,bai2025qwen25vl}. Given an input image,
the vision encoder extracts patch-level visual features, which are projected
into the language embedding space as visual tokens and combined with the
textual tokens. The resulting multimodal sequence is then processed through
stacked Transformer layers, where visual and textual information interact via
self-attention. The final hidden representations are used for multimodal
understanding and autoregressive generation.

\subsection{Token Scores Induce Spatial Allocation}
\label{sec:importance-criteria-spatial-allocations}

Most visual token pruning methods score individual tokens using signals such
as attention, relevance, saliency, or internal model responses, and retain the
highest-scoring tokens through global Top-$B$ selection. These methods focus
primarily on which tokens are more important, with less attention to how the
retained tokens are distributed in image space. We therefore ask: how does the
choice of scoring criterion shape the spatial allocation of the retained token
budget?

To answer this question, we conduct the analysis on
Qwen2.5-VL-7B~\cite{bai2025qwen25vl}. For each comparison, we keep the input,
token budget, and Top-$B$ selection rule fixed, varying only the scoring
criterion used for ranking. As shown in
\cref{fig:importance-spatial-allocations}(a), different scores produce
clearly different spatial layouts even for the same image. More importantly,
these differences remain when token selection frequencies are aggregated over
many distinct image--question pairs, as shown in
\cref{fig:importance-spatial-allocations}(b). At a coarse spatial scale,
FastV~\cite{chen2024fastv} and FEATHER no-RoPE~\cite{endo2025feather} both show
a clear preference for image boundaries: nearly $58\%$ of their retained
tokens lie in the edge region, substantially above the approximately $30.6\%$
expected from regional area. By comparison, VisionZip~\cite{yang2025visionzip}
and ERC exhibit more balanced edge--center distributions.

The fine-grained frequency maps show a more pronounced pattern. Several
scoring criteria produce persistent horizontal and vertical bands, together
with periodic row-and-column structures. Because these statistics aggregate
many images with different content and questions, object locations in
individual samples should be largely averaged out. Nevertheless, some fixed
token rows and columns continue to exhibit higher selection frequencies. This
finding indicates a stable positional dependence in token selection: the
resulting spatial distribution is not determined entirely by the current image
content or task semantics.

This phenomenon is not confined to a particular pruning layer. The scoring
criteria in \cref{fig:importance-spatial-allocations} operate either before the
visual tokens enter the LLM or at different early decoder layers, but all
exhibit stable spatial patterns to varying degrees. Moreover, FEATHER retains
a clear boundary preference at Layer~3 even after RoPE is removed. Delaying the
same no-RoPE criterion to Layer~8 does not eliminate this pattern. We therefore
do not attribute the positional dependence to any single pruning stage or
positional encoding, but regard it as a recurring spatial selection behavior
across different scoring criteria.

These results indicate that token importance alone does not fully determine
how tokens should be retained in space. Even when a scoring
criterion identifies high-scoring tokens, the resulting token distribution
may still exhibit a pronounced positional preference and may not adequately
reflect the spatial structure of the current image. Visual token pruning must
therefore consider the spatial organization of the retained tokens in addition
to the importance of each individual token.

\begin{figure}[t!]
  \centering
  \includegraphics[width=1\linewidth]{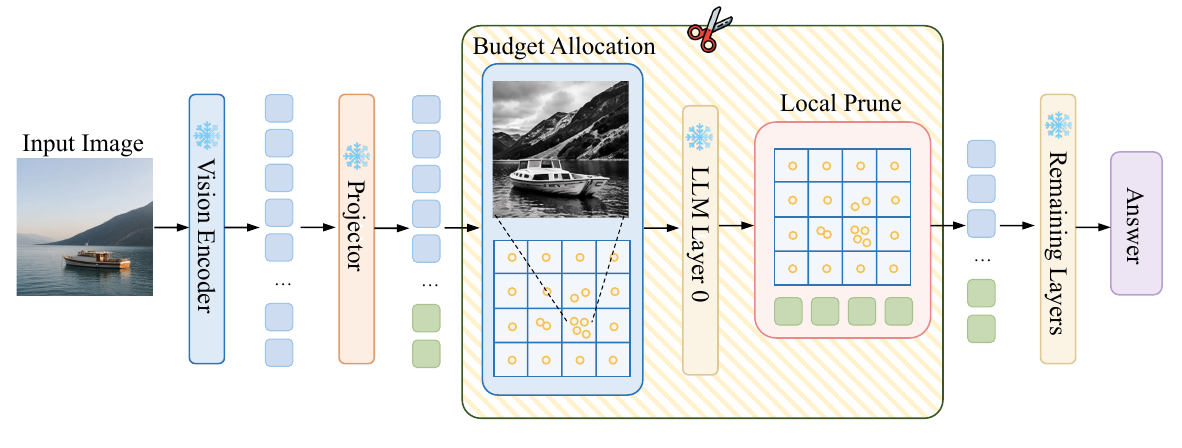}
  \caption{Overview of \methodname. Regional image structure determines the spatial scaffold, after which early decoder responses select local representatives under the prescribed token budget.}
  \label{fig:framework}
\end{figure}
\subsection{Broad Spatial Coverage Stabilizes Aggressive Pruning}
\label{sec:spatial-regularity}

\Cref{sec:importance-criteria-spatial-allocations} shows that different
selection signals leave stable but distinct positional preferences in sparse
visual representations. This makes it difficult to separate the quality of a
token score from the spatial organization of the resulting retained
representation. To examine their respective roles, we introduce two controls
that do not rely on sophisticated scoring signals: Random sampling and Uniform
Grid. These controls are not pruning methods proposed in this work. Random
sampling weakens criterion-specific positional preferences, whereas Uniform
Grid explicitly stabilizes spatial coverage; their comparison is shown in
\cref{fig:uniform-vs-baselines}.

The comparison presents a counterintuitive result: Random
sampling is already competitive with several carefully designed selection
criteria, and some methods fall below Random on multiple benchmarks. This
observation agrees with prior visual token studies that report strong random
baselines~\cite{wen2025rightproblem,wang2026random}. Because Random uses neither
token-level importance, attention, nor similarity, the result shows that a more
sophisticated scoring criterion does not automatically produce a more effective
sparse visual representation.

Uniform Grid further constrains the spatial coverage of visual tokens. Unlike
the sample-dependent random locations produced by Random, it distributes the
retained tokens consistently across different image regions.
\Cref{fig:uniform-vs-baselines} shows that, under the same 64-token budget,
Uniform Grid further outperforms Random on four of the five benchmarks and
remains competitive with complex pruning criteria on several tasks. Combined
with the position-dependent patterns in
\cref{sec:importance-criteria-spatial-allocations}, this result indicates
that the spatial coverage of the retained representation is not an incidental
by-product of token selection. A simple, stable spatial constraint can itself
provide strong pruning robustness. This trend is also consistent with
FEATHER~\cite{endo2025feather}, which uses uniform samples to maintain broad
image coverage.

The spatial coverage provided by Uniform Grid, however, is based on a fixed
sampling density. Smooth backgrounds and regions containing dense boundaries,
text, or fine-grained textures receive approximately the same local
representation resolution. Together, the Random and Uniform controls provide
a more specific design cue: stable spatial coverage is beneficial, but
coverage does not require the same representation density everywhere. We
therefore retain the broad spatial support provided by Uniform Grid while using
the local structure of the image to adapt its spatial resolution.
\subsection{Adaptive Spatial Density under Heterogeneous Structure}
\begin{wrapfigure}{r}{0.40\columnwidth}
  \vspace{-0.8\baselineskip}
  \centering
  \includegraphics[width=\linewidth]{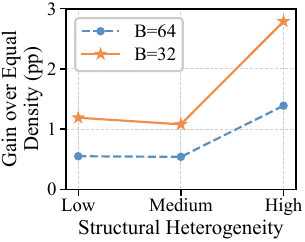}
  \caption{Gain of Adaptive over Equal Density on Qwen2.5-VL-7B, averaged over
  \datasetMMBEN{}, \datasetVQAText{}, and POPE. Gains generally increase with
  structural heterogeneity, especially at $B=32$.}
  \label{fig:heterogeneity}
  \vspace{-0.5\baselineskip}
\end{wrapfigure}
Uniform Grid preserves broad spatial coverage, but assigns approximately the
same representation density to different image regions. We next ask when this
equal-density assumption becomes restrictive. Intuitively, if local visual
structure is distributed relatively evenly, uniform sampling should already
provide a reasonable allocation. In contrast, when fine-grained structure is
concentrated in only a few regions, spending the same number of tokens
everywhere may leave these regions underrepresented.

To isolate this factor, we conduct a controlled allocation analysis in which
the token budget and local representative rule are kept identical. Equal
Density assigns approximately the same number of representation slots to each
coarse region, whereas Adaptive Density varies the regional budgets according
to local structural variation. Neither setting uses semantic relevance,
attention, or decoder-derived token scores. We quantify how unevenly local
structure is distributed across an image by

\begin{equation}
    H_{\mathrm{str}}
=
\frac{\mathrm{Std}(\{c_g\}_{g=1}^{G})}
     {\mathrm{Mean}(\{c_g\}_{g=1}^{G})+\epsilon},
\
\end{equation}

where $c_g$ denotes the structural variation of region $g$. Samples are grouped
into low-, medium-, and high-heterogeneity subsets according to
$H_{\mathrm{str}}$.

As shown in Fig.~\ref{fig:heterogeneity}, adaptive density provides only a
modest advantage when regional structure is relatively homogeneous, whereas
the gap becomes substantially larger for images with highly heterogeneous
structure. This effect is more pronounced under the tighter $B=32$ budget.
Averaged across MMBench, TextVQA, and POPE, the high-heterogeneity group gains
1.39 points at $B=64$ and 2.79 points at $B=32$, clearly exceeding the
corresponding gains for the low- and medium-heterogeneity groups. The trend is
not strictly monotonic on every individual benchmark, indicating that
structural heterogeneity alone does not determine pruning difficulty.

These observations show a limitation of uniform spatial sampling: the issue
is not broad coverage itself, but the use of a fixed representation density
everywhere. When local structure is distributed evenly, equal density already
provides a strong sparse representation. When structure is highly uneven,
however, uniformly spreading a limited number of tokens can over-represent
simple regions while leaving detail-rich regions under-represented. This
mismatch becomes more consequential as the token budget shrinks. These findings
motivate preserving broad spatial support while adapting local token density
to the structure of each image.
\section{Method}
\label{sec:laplacian-guided-allocation}

Based on the preceding analysis, we propose \methodname, a training-free method
that preserves broad spatial support while adapting representation density to
regional structure. It first converts the global token budget into
region-specific capacities and local cells using Laplacian variation
(\cref{sec:regional-capacity}). It then uses Early Representation Change
(ERC) after the first decoder block to select one representative from each cell
(\cref{sec:local-representative-selection}). These stages separate where
the visual budget is spent from which token represents each cell, leaving
exactly $B$ tokens for all remaining decoder layers. The overall framework is
shown in \cref{fig:framework}.

\begin{table*}[t]
  \centering
  \caption{Main results on Qwen2.5-VL-7B under different token budgets. We compare FastV~\cite{chen2024fastv}, DART~\cite{wen2025dart}, GPrune~\cite{jiang2025gprune}, VisionZip~\cite{yang2025visionzip}, VisPruner~\cite{zhang2025vispruner}, GridPrune~\cite{duan2025gridprune}, and HoloV~\cite{zou2025holov}. Acc. averages all ten benchmark scores after dividing MME by 28. Rel. first computes each pruned-to-full score ratio and then averages the ten ratios. Bold entries are the best results within each budget, and underlined entries are the second best.}
  \label{tab:main-results}

  \small
  \setlength{\tabcolsep}{1.2pt}
  \renewcommand{\arraystretch}{1.0}

  \newcommand{\bestresult}[1]{\textbf{#1}}
  \newcommand{\secondresult}[1]{\underline{#1}}

  \resizebox{\textwidth}{!}{%
  \begin{tabular}{l|cccccccccc|cc}
    \toprule
    \textbf{Method} &
    \textbf{GQA} &
    \textbf{\datasetSQA} &
    \textbf{\datasetVQAText} &
    \textbf{VizWiz} &
    \textbf{MMMU} &
    \textbf{POPE} &
    \textbf{MME} &
    \textbf{\datasetMMBEN} &
    \textbf{\datasetMMBCN} &
    \textbf{MMVet} &
    \textbf{Acc.} &
    \textbf{Rel.} \\
    \midrule

    \rowcolor{resultgroupgray}
    \multicolumn{13}{c}{\textit{Upper Bound, All 576 Tokens (100\%)}} \\
    \rowcolor{resultuppergray}
    Qwen2.5-VL-7B~\cite{bai2025qwen25vl} & 57.4 & 87.6 & 80.8 & 36.8 & 51.7 & 83.9 & 2347 & 88.7 & 86.9 & 31.7 & 68.9 & 100.0\% \\

    \rowcolor{resultgroupgray}
    \multicolumn{13}{c}{\textit{Retain 128 Tokens} \textcolor{resultreduction}{($\downarrow 77.8\%$)}} \\
    \rowcolor{resultattention}
    FastV (ECCV24)~\cite{chen2024fastv} & 51.3 & 83.4 & \bestresult{66.3} & \secondresult{34.1} & \bestresult{51.2} & 79.5 & 2093 & 84.5 & 83.9 & 24.8 & 63.4 & 91.2\% \\
    \rowcolor{resultadaptive}
    DART (EMNLP25)~\cite{wen2025dart} & 41.8 & 79.8 & 55.3 & 29.6 & 47.1 & 42.1 & 2053 & 84.4 & 84.0 & 24.8 & 56.2 & 81.1\% \\
    \rowcolor{resultmerge}
    GPrune (AAAI25)~\cite{jiang2025gprune} & 50.5 & 82.1 & 48.9 & 26.5 & 46.4 & 77.0 & 1771 & 81.7 & 82.0 & 19.7 & 57.8 & 82.0\% \\
    \rowcolor{resultmerge}
    VisionZip (CVPR25)~\cite{yang2025visionzip} & 52.8 & 84.4 & 54.0 & 33.3 & 48.8 & 78.0 & 2071 & 85.1 & 84.7 & 23.4 & 61.8 & 88.8\% \\
    \rowcolor{resultmerge}
    VisPruner (ICCV25)~\cite{zhang2025vispruner} & 53.6 & \bestresult{85.5} & 63.6 & 33.4 & 49.1 & 79.8 & 2072 & \secondresult{85.7} & \bestresult{85.9} & 23.4 & 63.4 & 90.8\% \\
    \rowcolor{resultmerge}
    GridPrune (arXiv25)~\cite{duan2025gridprune} & \secondresult{54.4} & 84.8 & 61.2 & \secondresult{34.1} & 48.2 & 79.5 & 2101.9 & 85.3 & 85.3 & \bestresult{26.2} & 63.4 & 91.5\% \\
    \rowcolor{resultmerge}
    HoloV (NeurIPS25)~\cite{zou2025holov} & 53.6 & 84.4 & \secondresult{64.4} & \bestresult{34.2} & 47.9 & \bestresult{81.0} & \bestresult{2178.3} & 85.5 & 85.1 & \secondresult{25.7} & \secondresult{63.9} & \secondresult{92.0\%} \\
    \rowcolor{resultours}
    \textbf{\methodname (Ours)} & \bestresult{55.3} & \secondresult{85.4} & 62.3 & 33.5 & \secondresult{49.3} & \secondresult{80.1} & \secondresult{2168} & \bestresult{86.2} & \secondresult{85.7} & 25.2 & \bestresult{64.0} & \bestresult{92.1\%} \\

    \rowcolor{resultgroupgray}
    \multicolumn{13}{c}{\textit{Retain 64 Tokens} \textcolor{resultreduction}{($\downarrow 88.9\%$)}} \\
    \rowcolor{resultattention}
    FastV (ECCV24)~\cite{chen2024fastv} & 44.9 & 78.7 & \bestresult{56.6} & 29.1 & 46.8 & 68.9 & 1735 & 74.7 & 74.1 & 16.5 & 55.2 & 78.5\% \\
    \rowcolor{resultadaptive}
    DART (EMNLP25)~\cite{wen2025dart} & 42.9 & 78.9 & 42.8 & 26.0 & 46.2 & 69.5 & 1936 & 82.4 & 81.7 & 21.1 & 56.1 & 79.7\% \\
    \rowcolor{resultmerge}
    GPrune (AAAI25)~\cite{jiang2025gprune} & 44.6 & 79.7 & 34.2 & 19.2 & 42.8 & 67.2 & 1427 & 74.4 & 74.9 & 16.1 & 50.4 & 70.8\% \\
    \rowcolor{resultmerge}
    VisionZip (CVPR25)~\cite{yang2025visionzip} & \secondresult{50.7} & 80.6 & 40.2 & 29.0 & 45.8 & 72.5 & 1885 & 83.2 & 82.8 & 21.1 & 57.3 & 82.0\% \\
    \rowcolor{resultmerge}
    VisPruner (ICCV25)~\cite{zhang2025vispruner} & 50.5 & 81.6 & 50.8 & 29.3 & \secondresult{47.2} & 74.6 & 1850 & 83.1 & 83.2 & 22.5 & 58.9 & 84.3\% \\
    \rowcolor{resultmerge}
    GridPrune (arXiv25)~\cite{duan2025gridprune} & 49.7 & \secondresult{82.1} & 52.8 & \secondresult{30.6} & \bestresult{47.6} & 71.2 & 1865.6 & 83.6 & 83.1 & 21.6 & 58.9 & 84.3\% \\
    \rowcolor{resultmerge}
    HoloV (NeurIPS25)~\cite{zou2025holov} & 50.2 & \bestresult{83.9} & 52.6 & 30.3 & 45.4 & \bestresult{76.6} & \secondresult{2022.6} & \secondresult{84.1} & \bestresult{84.4} & \secondresult{22.9} & \secondresult{60.3} & \secondresult{86.0\%} \\
    \rowcolor{resultours}
    \textbf{\methodname (Ours)} & \bestresult{52.5} & 81.0 & \secondresult{55.2} & \bestresult{31.0} & 46.3 & \secondresult{74.8} & \bestresult{2035} & \bestresult{84.6} & \secondresult{84.3} & \bestresult{23.9} & \bestresult{60.6} & \bestresult{87.0\%} \\

    \rowcolor{resultgroupgray}
    \multicolumn{13}{c}{\textit{Retain 32 Tokens} \textcolor{resultreduction}{($\downarrow 94.4\%$)}} \\
    \rowcolor{resultattention}
    FastV (ECCV24)~\cite{chen2024fastv} & 37.6 & 73.2 & 34.1 & 20.3 & 42.7 & 39.1 & 1222 & 58.2 & 57.9 & 7.3 & 41.4 & 58.3\% \\
    \rowcolor{resultadaptive}
    DART (EMNLP25)~\cite{wen2025dart} & 45.4 & 76.4 & 33.4 & 22.4 & 43.0 & 58.1 & 1696 & 78.5 & 78.8 & 17.4 & 51.4 & 72.7\% \\
    \rowcolor{resultmerge}
    GPrune (AAAI25)~\cite{jiang2025gprune} & 37.9 & 76.6 & 22.0 & 13.8 & 42.4 & 48.8 & 1187 & 66.9 & 67.7 & 12.4 & 43.1 & 60.1\% \\
    \rowcolor{resultmerge}
    VisionZip (CVPR25)~\cite{yang2025visionzip} & \secondresult{46.4} & 76.1 & 28.8 & 25.4 & \secondresult{45.8} & 63.2 & 1672 & 80.4 & 80.1 & 17.4 & 52.3 & 74.5\% \\
    \rowcolor{resultmerge}
    VisPruner (ICCV25)~\cite{zhang2025vispruner} & 44.2 & 78.5 & 36.4 & 23.8 & 44.4 & 63.2 & 1560 & 78.6 & 79.7 & 17.9 & 52.2 & 74.1\% \\
    \rowcolor{resultmerge}
    GridPrune (arXiv25)~\cite{duan2025gridprune} & 45.4 & \secondresult{80.1} & \secondresult{43.8} & \secondresult{27.2} & 45.4 & 61.6 & 1646.1 & 80.2 & 80.3 & \bestresult{20.6} & 54.4 & 77.8\% \\
    \rowcolor{resultmerge}
    HoloV (NeurIPS25)~\cite{zou2025holov} & 45.1 & \bestresult{80.8} & 40.4 & 25.8 & \bestresult{46.4} & \bestresult{69.7} & \secondresult{1745.7} & \secondresult{81.0} & \secondresult{80.6} & \secondresult{18.4} & \secondresult{55.0} & \secondresult{78.0\%} \\
    \rowcolor{resultours}
    \textbf{\methodname (Ours)} & \bestresult{49.0} & 78.4 & \bestresult{44.9} & \bestresult{27.8} & 43.0 & \secondresult{66.9} & \bestresult{1844} & \bestresult{82.6} & \bestresult{81.8} & \secondresult{18.4} & \bestresult{55.9} & \bestresult{79.3\%} \\

    \bottomrule
  \end{tabular}%
  }
\end{table*}

\newcommand{\crossarchitecturetable}{%
  \setlength{\tabcolsep}{4.0pt}
  \renewcommand{\arraystretch}{1}
  \resizebox{\columnwidth}{!}{%
  \begin{tabular}{l|cccc|cc}
    \toprule
    \textbf{Method} &
    \textbf{\datasetSQA} &
    \textbf{\datasetMMBEN} &
    \textbf{MMMU} &
    \textbf{OCRBench} &
    \textbf{Acc.} &
    \textbf{Rel.} \\
    \midrule

    \rowcolor{resultgroupgray}
    \multicolumn{7}{c}{\textit{Upper Bound, All 729 Tokens (100\%)}} \\
    \rowcolor{resultuppergray}
    LLaVA-OneVision-7B & 91.3 & 85.4 & 45.3 & 49.7 & 67.9 & 100.0\% \\

    \rowcolor{resultgroupgray}
    \multicolumn{7}{c}{\textit{Retain 160 Tokens} \textcolor{resultreduction}{($\downarrow 78.1\%$)}} \\
    \rowcolor{resultattention}
    FastV~\cite{chen2024fastv} & 80.7 & 79.7 & \underline{44.3} & 24.9 & 57.4 & 82.4\% \\
    \rowcolor{resultmerge}
    CDPruner~\cite{zhang2025cdpruner} & \underline{83.6} & 80.1 & 42.8 & 34.7 & 60.3 & 87.4\% \\
    \rowcolor{resultmerge}
    GridPrune~\cite{duan2025gridprune} & 82.5 & \underline{81.5} & 44.1 & \underline{35.2} & \underline{60.8} & \underline{88.5\%} \\
    \rowcolor{resultours}
    \textbf{\methodname (Ours)} & \textbf{87.5} & \textbf{82.8} & \textbf{46.0} & \textbf{41.7} & \textbf{64.5} & \textbf{94.6\%} \\

    \rowcolor{resultgroupgray}
    \multicolumn{7}{c}{\textit{Retain 80 Tokens} \textcolor{resultreduction}{($\downarrow 89.0\%$)}} \\
    \rowcolor{resultattention}
    FastV~\cite{chen2024fastv} & 77.2 & 75.7 & \underline{43.7} & 13.0 & 52.4 & 74.0\% \\
    \rowcolor{resultmerge}
    CDPruner~\cite{zhang2025cdpruner} & \underline{80.4} & \underline{76.7} & 42.9 & \underline{24.4} & \underline{56.1} & \underline{80.4\%} \\
    \rowcolor{resultmerge}
    GridPrune~\cite{duan2025gridprune} & 77.8 & 75.8 & \underline{43.7} & 22.5 & 55.0 & 78.9\% \\
    \rowcolor{resultours}
    \textbf{\methodname (Ours)} & \textbf{83.5} & \textbf{80.0} & \textbf{44.1} & \textbf{32.1} & \textbf{59.9} & \textbf{86.8\%} \\

    \rowcolor{resultgroupgray}
    \multicolumn{7}{c}{\textit{Retain 40 Tokens} \textcolor{resultreduction}{($\downarrow 94.5\%$)}} \\
    \rowcolor{resultattention}
    FastV~\cite{chen2024fastv} & 75.8 & 68.6 & \underline{42.0} & 8.0 & 48.6 & 68.0\% \\
    \rowcolor{resultmerge}
    CDPruner~\cite{zhang2025cdpruner} & \underline{78.5} & \underline{70.3} & 41.3 & \underline{16.9} & \underline{51.8} & \underline{73.4\%} \\
    \rowcolor{resultmerge}
    GridPrune~\cite{duan2025gridprune} & 76.0 & 69.6 & 41.9 & 16.0 & 50.9 & 72.4\% \\
    \rowcolor{resultours}
    \textbf{\methodname (Ours)} & \textbf{78.9} & \textbf{71.5} & \textbf{42.6} & \textbf{22.1} & \textbf{53.8} & \textbf{77.2\%} \\
    \bottomrule
  \end{tabular}%
  }
}

\subsection{Adaptive Spatial Scaffold}
\label{sec:regional-capacity}

Uniform sampling preserves image coverage but fixes the representation density
across space. Smooth regions can be represented sparsely, but boundaries, text,
thin structures, and fine textures often require denser sampling. Our scaffold
keeps broad spatial support and varies its local density.

We divide the visual-token grid into $G$ disjoint coarse regions
$\{\mathcal{R}_g\}_{g=1}^{G}$, with $G\leq B$. We first convert the complete
input image to grayscale and compute a four-neighbor Laplacian response map
$L$ using replicate padding at the outer image boundary. We then map each
token region $\mathcal{R}_g$ to its image-domain support $\Omega_g$ and crop
the already computed response map. Its raw structural score is
\begin{equation}
    c_g^{\mathrm{raw}} = \operatorname{Var}\!\left(L|_{\Omega_g}\right).
    \label{eq:laplacian-complexity}
\end{equation}
Thus, neighboring pixels across internal region boundaries contribute to the
response, while replicate padding is applied only at the outer image boundary.
The raw scores are independently min--max normalized within each image:
\begin{equation}
    c_g =
    \frac{c_g^{\mathrm{raw}}-\min_j c_j^{\mathrm{raw}}}
    {\max\!\left(\max_j c_j^{\mathrm{raw}}-\min_j c_j^{\mathrm{raw}},
    \epsilon_{\mathrm{norm}}\right)}.
    \label{eq:laplacian-normalization}
\end{equation}
The Laplacian responds weakly in smooth regions and strongly around rapid
local changes~\cite{marr1980edgedetection,burt1983laplacian}; in the frequency
domain, its response grows quadratically with spatial frequency. The
normalized score $c_g$ measures local structural variation rather than
semantic relevance to the question.

Each coarse region receives one token to preserve broad spatial support. We
allocate the remaining $B-G$ tokens in proportion to $c_g$, subject to the
regional capacities:
\begin{equation}
    \begin{aligned}
    w_g &= \frac{c_g}{\sum_{j=1}^{G}c_j}, \\
    1 \leq B_g &\leq |\mathcal{R}_g|,
    \qquad \sum_{g=1}^{G} B_g = B.
    \end{aligned}
    \label{eq:regional-budget}
\end{equation}
\Cref{eq:regional-budget} reserves one token per region and caps its budget by
the number of tokens it contains. If all structural scores are zero, we use
$w_g=1/G$. An iterative largest-remainder procedure distributes the remaining
budget among unsaturated regions. When a region reaches $|\mathcal{R}_g|$, its
excess allocation is redistributed until exactly $B$ tokens are assigned.
Each region $\mathcal{R}_g$ is then divided into $B_g$ cells
$\{\mathcal{C}_{g,k}\}_{k=1}^{B_g}$ of approximately equal area. Smooth regions
retain a sparse scaffold, whereas regions with more local variation are
represented at a higher spatial density.

\subsection{Local Selection with Early Decoder Responses}
\label{sec:local-representative-selection}

Once the spatial cells are fixed, we select tokens independently within each
cell. In a residual Transformer block~\cite{vaswani2017attention},
the update at position $i$ can be written as
$h_i^{l+1}=h_i^l+\Delta_i^l$. Its magnitude is
$r_i^{(l)}=\lVert h_i^{l+1}-h_i^l\rVert_2$. We use the response from the first
decoder block,
\begin{equation}
    r_i=r_i^{(0)}=\lVert h_i^1-h_i^0\rVert_2
    =\lVert\Delta_i^0\rVert_2,
    \label{eq:erc}
\end{equation}
We call the score in \cref{eq:erc} \emph{Early Representation Change} (ERC). ERC measures the update
produced by the decoder. It does not measure the magnitude of the incoming
visual feature.

For each cell $\mathcal{C}_{g,k}$, we retain
\begin{equation}
    i_{g,k}^{\star}
    =\arg\max_{i\in\mathcal{C}_{g,k}} r_i.
    \label{eq:local-representative}
\end{equation}
\Cref{eq:local-representative} applies the score only within a cell. We do not
compare it across distant regions. The scaffold has already
fixed where the budget is spent; ERC only chooses a representative within each
allocated cell.

We compute ERC after the first decoder block, where a decoder response first
becomes available. Pruning at a deeper layer would require processing the full
visual sequence through additional blocks. After Layer~0, we remove unselected
visual tokens and preserve the order and positional information of the retained
$B$ tokens in the remaining decoder layers.

\begin{table}[t]
  \centering
  \caption{Cross-architecture results on LLaVA-OneVision-7B. Acc./Rel. denote
  four-benchmark averages and relative retention; bold/underlined mark the best
  and second-best results.}
  \label{tab:cross-architecture}
  \crossarchitecturetable
\end{table}

\section{Experiments}
\label{sec:experiments}

We evaluate our method on ten multimodal understanding benchmarks at several
visual token budgets and report comparisons, ablations, and inference cost.

\subsection{Experimental Setup}
\label{sec:experimental-setup}

\paragraph{Datasets.}
We use ten image benchmarks. MMMU~\cite{yue2024mmmu} tests expert multimodal
reasoning; GQA~\cite{hudson2019gqa}, compositional visual reasoning;
\datasetSQA~\cite{lu2022scienceqa}, multimodal science question answering; and
\datasetVQAText~\cite{singh2019textvqa}, reading text in images.
VizWiz~\cite{gurari2018vizwiz} covers visual questions submitted by blind users;
POPE~\cite{li2023pope} evaluates object hallucination;
MME~\cite{fu2023mme}, perception and cognition;
\datasetMMBEN{} and \datasetMMBCN{}~\cite{liu2023mmbench}, general multimodal
understanding in English and Chinese; and
MMVet~\cite{yu2024mmvet}, integrated multimodal capabilities. We follow the
official evaluation protocol and metric for each benchmark.

\paragraph{Model architectures.}
We use Qwen2.5-VL-7B-Instruct~\cite{bai2025qwen25vl} for the main benchmark
comparisons and ablations, and LLaVA-OneVision-7B for cross-architecture
evaluation. Unless stated otherwise, all analyses use Qwen2.5-VL-7B-Instruct.

\paragraph{Baselines.}
We compare with seven recent methods that require no training. FastV uses
attention~\cite{chen2024fastv}. DART removes duplicated tokens~\cite{wen2025dart},
whereas GPrune selects representative tokens through graph-based information
propagation~\cite{jiang2025gprune}. VisionZip combines visual attention with
similarity-based merging~\cite{yang2025visionzip}, and VisPruner combines visual
attention with diversity~\cite{zhang2025vispruner}. GridPrune performs
query-conditioned regional budget allocation followed by local token
selection~\cite{duan2025gridprune}, while HoloV distributes the pruning budget
across spatial crops to retain holistic visual context~\cite{zou2025holov}.

\paragraph{Implementation details.}
Qwen2.5-VL produces 576 post-merge visual tokens on a $24\times24$ grid. We
evaluate $B\in\{128,64,32\}$ (reductions of $77.8\%$, $88.9\%$, and $94.4\%$)
using $8\times8$, $5\times5$, and $4\times4$ coarse grids, respectively. For
each grayscale image, we compute one full-image four-neighbor Laplacian
response map with replicate padding, crop the mapped regional responses, and
min--max normalize their variances within the image. We allocate budgets in
proportion to these normalized scores and use largest-remainder rounding to obtain integer $B_g$
with $1\leq B_g\leq|\mathcal{R}_g|$ and $\sum_g B_g=B$. Each $\mathcal{R}_g$ is divided into $B_g$ approximately
equal-area subcells, each retaining the token with the largest Early
Representation Change (ERC). Pruning follows decoder Layer~0 and occurs before
Layer~1.

\subsection{Main Results}
\label{sec:main-results}

\begin{figure*}[t]
  \centering
  \includegraphics[width=0.84\textwidth]{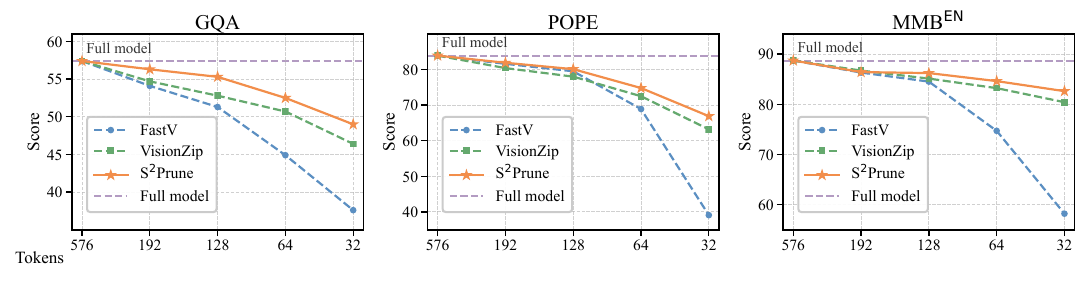}
  \caption{Qwen2.5-VL-7B performance across retained visual-token budgets on
  GQA, POPE, and \datasetMMBEN. Dashed lines denote the unpruned 576-token
  model.}
  \label{fig:token-budget-performance}
\end{figure*}

\Cref{tab:main-results} compares \methodname with seven pruning approaches
that require no training on Qwen2.5-VL-7B. Our method has the highest aggregate
accuracy with 128, 64, and 32 visual tokens. At 128 tokens, it reaches 64.0
Acc. and retains 92.1\% of the full-model performance. At 64 tokens, it reaches
60.6 Acc. and 87.0\% relative performance. Across all three budgets, our method
gives the best GQA and \datasetMMBEN{} results. On
\datasetMMBCN, it ranks second at 128 tokens and first at
64 and 32 tokens. At
32 tokens, it reaches 55.9 Acc. and 79.3\% relative performance, leading on
six of the ten benchmarks, including VizWiz and \datasetMMBCN. The
corresponding Acc. margins over the strongest baseline are 0.1, 0.3, and 0.9
points at 128, 64, and 32 tokens, respectively.
\Cref{fig:token-budget-performance} compares performance as the
visual-token budget decreases from 576 to 32. Among the three methods shown,
\methodname achieves the highest scores on GQA and POPE at every pruned budget
and on \datasetMMBEN{} at 128, 64, and 32 tokens; at 192 tokens, it remains
within 0.3 points of VisionZip on \datasetMMBEN. At 32 tokens, its gains over
the stronger of the two plotted baselines are 2.6, 3.7, and 2.2 points on GQA,
POPE, and \datasetMMBEN, respectively. FastV degrades sharply on POPE and
\datasetMMBEN, whereas VisionZip is more stable; \methodname nevertheless
achieves the highest score on all three benchmarks at this lowest budget. This
trend suggests that
structure-adaptive spatial allocation better preserves broad spatial support
and local detail when few tokens remain.

\subsection{\texorpdfstring{\methodname}{S2Prune} with Another VLM Architecture}
\label{sec:cross-architecture}

\Cref{tab:cross-architecture} evaluates \methodname on
LLaVA-OneVision-7B at budgets of 160, 80, and 40 tokens. \methodname leads all
four benchmarks and attains the highest Acc. and Rel. at every budget. These
results support the transferability of the pruning strategy beyond Qwen2.5-VL.

\subsection{Efficiency Analysis}
\label{sec:efficiency}

\begin{table}[t]
  \centering
  \caption{Efficiency on Qwen2.5-VL-7B at $B=64$. Acc. is POPE classification
  accuracy (the main results use POPE F1); runtime is also measured on POPE.}
  \label{tab:efficiency}
  \footnotesize
  \setlength{\tabcolsep}{2.0pt}
  \renewcommand{\arraystretch}{1}
  \resizebox{0.8\columnwidth}{!}{%
  \begin{tabular}{lcccc}
    \toprule
    \textbf{Method} &
    \textbf{Acc. $\uparrow$} &
    \shortstack{\textbf{FLOPs}\\\textbf{(T) $\downarrow$}} &
    \shortstack{\textbf{Prefill}\\\textbf{(ms) $\downarrow$}} &
    \shortstack{\textbf{KV Cache}\\\textbf{(MB) $\downarrow$}} \\
    \midrule
    \rowcolor{resultuppergray}
    Full & 71.3 & 8.181 & 56.97 & 31.50 \\
    \midrule
    FastV & 57.7 & 1.839 & \underline{24.86} & 5.50 \\
    GPrune & 51.6 & \textbf{1.352} & \textbf{20.49} & \textbf{3.50} \\
    VisPruner & \underline{60.4} & \textbf{1.352} & 26.24 & \textbf{3.50} \\
    \rowcolor{resultours}
    \textbf{\methodname (Ours)} & \textbf{62.3} & \underline{1.596} & 28.66 & \underline{4.50} \\
    \bottomrule
  \end{tabular}
  }
\end{table}

To demonstrate the efficiency of \methodname, we compare it with representative
visual token pruning methods in terms of FLOPs, prefill latency, and KV-cache
consumption on Qwen2.5-VL-7B. All experiments are performed on a single NVIDIA
GeForce RTX 5090 GPU. We use POPE for the efficiency evaluation because its
questions have similar lengths and each sample involves one prefill stage and
one decoding stage. As shown in \cref{tab:efficiency}, when the number of
visual tokens is reduced from 576 to 64, \methodname reduces FLOPs from 8.181T
to 1.596T, corresponding to a 5.1$\times$ reduction. It also reduces prefill
latency from 56.97 ms to 28.66 ms, which is nearly a 2.0$\times$ speedup, and
decreases the KV cache from 31.50 MB to 4.50 MB, corresponding to a
7.0$\times$ reduction. Although GPrune
and VisPruner achieve lower computational costs, \methodname maintains the
highest average accuracy among the compared pruning methods, demonstrating a
favorable trade-off between inference efficiency and model performance.

\subsection{Ablation Study}
\label{sec:ablation-study}

\begin{figure}[t]
  \centering
  \includegraphics[width=0.98\columnwidth]{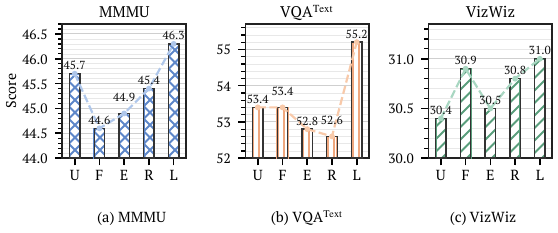}
  \caption{Structural-prior ablation at $B=64$. U/F/E/R/L denote Uniform, FFT,
  Effective Rank, Random, and Laplacian; Laplacian is best on all three tasks.}
  \label{fig:token-source-ablation}
\end{figure}

\paragraph{Effect of Structural Prior.}
We fix the spatial partition, 64-token budget, and local ERC selector, and vary
only the regional allocation strategy. Uniform assigns the same capacity to
each cell, whereas Random distributes the fixed budget randomly across cells.
We also compare frequency-domain FFT energy, Effective Rank from visual
features~\cite{roy2007effectiverank}, and Laplacian variation as data-dependent
capacity signals. \Cref{fig:token-source-ablation} shows that Laplacian
gives the highest scores on all three benchmarks, reaching 46.3 on MMMU, 55.2
on \datasetVQAText, and 31.0 on VizWiz. The strongest alternatives reach 45.7,
53.4, and 30.9, respectively. Because the partition, budget, and local ERC
selector remain fixed, these differences isolate the regional allocation
strategy.

\paragraph{Effect of Local Representative Selection.}
With spatial allocation and pruning fixed, we compare Random, Hidden Norm
($\lVert h_i^1\rVert_2$), Query Attn. using all question tokens or the last
valid token, and ERC ($\lVert h_i^1-h_i^0\rVert_2$).
\Cref{tab:local-selector-ablation} shows that ERC leads all three tasks
(84.6/81.0/55.2), producing the best Acc./Rel. (73.6/85.4\%). Query Attn.
(Last Token) ranks second in aggregate performance (72.3/83.9\%). ERC is
therefore the most consistent local selector under the same spatial allocation.

\begin{table}[t]
  \centering
  \caption{Local-selector ablation at $B=64$ on a $5\times5$ grid. Acc./Rel.
  denote three-benchmark averages and relative retention.}
  \label{tab:local-selector-ablation}
  \footnotesize
  \setlength{\tabcolsep}{2.5pt}
  \renewcommand{\arraystretch}{0.94}
  \resizebox{0.98\columnwidth}{!}{%
  \begin{tabular}{lccc|cc}
    \toprule
    \textbf{Local Selector} &
    \textbf{\datasetMMBEN} &
    \textbf{\datasetSQA} &
    \textbf{\datasetVQAText} &
    \textbf{Acc.} &
    \textbf{Rel.} \\
    \midrule
    Random & \underline{83.5} & 79.6 & 44.9 & 69.3 & 80.2\% \\
    Hidden Norm & 82.8 & \underline{80.8} & 30.5 & 64.7 & 74.4\% \\
    Query Attn. (All-Q) & 82.8 & 80.2 & 52.4 & 71.8 & 83.3\% \\
    Query Attn. (Last Token) & 82.7 & 80.4 & \underline{53.8} & \underline{72.3} & \underline{83.9\%} \\
    \rowcolor{resultours}
    \textbf{ERC (Ours)} & \textbf{84.6} & \textbf{81.0} & \textbf{55.2} & \textbf{73.6} & \textbf{85.4\%} \\
    \bottomrule
  \end{tabular}%
  }
\end{table}

\section{Conclusion}
\label{sec:conclusion}

In this paper, we have revisited visual token pruning from a spatial allocation perspective.
We show that different token-selection criteria induce stable spatial biases
and do not always outperform Uniform Grid sampling under aggressive pruning.
Based on this observation, we propose \methodname, a training-free method that
separates regional budget allocation from local token selection. \methodname
uses the variance of Laplacian responses to measure regional structural
complexity and allocate token budgets, while ERC selects representative tokens
using the first decoder response. Experiments across ten benchmarks and two
MLLM architectures show consistent gains over training-free baselines. These results highlight that effective visual token pruning should consider both \emph{which} tokens are selected and \emph{where}
the token budget is spent.

\FloatBarrier
{
    \fontsize{9}{9.8}\selectfont
    \setlength{\bibsep}{0pt}
    \bibliographystyle{ieeenat_fullname}
    \bibliography{main}
}

\clearpage
\appendix
\setcounter{figure}{0}
\setcounter{table}{0}
\setcounter{equation}{0}
\setcounter{algorithm}{0}
\renewcommand{\thefigure}{S\arabic{figure}}
\renewcommand{\thetable}{S\arabic{table}}
\renewcommand{\theequation}{S\arabic{equation}}
\renewcommand{\thealgorithm}{S\arabic{algorithm}}
\renewcommand{\theHfigure}{supp.figure.\arabic{figure}}
\renewcommand{\theHtable}{supp.table.\arabic{table}}
\renewcommand{\theHequation}{supp.equation.\arabic{equation}}
\renewcommand{\theHalgorithm}{supp.algorithm.\arabic{algorithm}}
\makeatletter
\setlength{\@fptop}{0pt}
\setlength{\@dblfptop}{0pt}
\makeatother
\def\SUPPLEMENTINMAIN{}

\ifdefined\SUPPLEMENTINMAIN
\else

\documentclass[10pt,twocolumn,letterpaper]{article}

\usepackage[pagenumbers]{paperstyle}
%
%
\newcommand{\red}[1]{{\color{red}#1}}
\newcommand{\todo}[1]{{\color{red}#1}}
\newcommand{\TODO}[1]{\textbf{\color{red}[TODO: #1]}}

\usepackage{colortbl}
\definecolor{resultgroupgray}{HTML}{E4E4E4}
\definecolor{resultuppergray}{HTML}{FFFFFF}
\definecolor{resultattention}{HTML}{FFFFFF}
\definecolor{resultmerge}{HTML}{FFFFFF}
\definecolor{resultadaptive}{HTML}{FFFFFF}
\definecolor{resultours}{HTML}{F1ECF7}
\definecolor{resultreduction}{HTML}{74569B}

\makeatletter
\IfFormatAtLeastTF{2025-06-01}{%
  \@ifpackageloaded{lineno}{%
    \RemoveFromHook{build/column/before}[lineno]%
    \AddToHook{build/column/before}[paper-lineno-column-fix]{%
      \if@twocolumn
        \if@firstcolumn
          \protected@write\@auxout{}{\string\def\string\@LN@column{2}}%
        \else
          \protected@write\@auxout{}{\string\def\string\@LN@column{1}}%
        \fi
      \fi
    }%
  }{}%
}{}%
\makeatother

\definecolor{resultours}{HTML}{FFF1E6}
\definecolor{resultreduction}{HTML}{D9772E}
\usepackage{float}
\usepackage{placeins}
\usepackage{algorithm}
\usepackage[noend]{algpseudocode}
\usepackage{cuted}

\makeatletter
\setlength{\@fptop}{0pt}
\setlength{\@dblfptop}{0pt}
\makeatother

\definecolor{linkblue}{rgb}{0.21,0.49,0.74}
\usepackage[pagebackref,breaklinks,colorlinks,allcolors=linkblue]{hyperref}
\makeatletter
\def\theHALG@line{\thealgorithm.\arabic{ALG@line}}
\makeatother

\newcommand{\methodname}{S\textsuperscript{2}Prune\xspace}
\newcommand{\datasetSQA}{SQA\textsuperscript{IMG}}
\newcommand{\datasetVQAText}{VQA\textsuperscript{Text}}
\newcommand{\datasetMMBEN}{MMB\textsuperscript{EN}}
\newcommand{\datasetMMBCN}{MMB\textsuperscript{CN}}

\algrenewcommand\algorithmicrequire{\textbf{Input:}}
\algrenewcommand\algorithmicensure{\textbf{Output:}}

\title{S\textsuperscript{2}Prune: Spatially Structured Visual Token Pruning for Multimodal Large Language Models\\[0.35em]
{\large Supplementary Material}}
\author{Yuanyuan Jia \qquad Shunpu Tang \qquad Qianqian Yang\\
College of Information Science and Electronic Engineering, Zhejiang University\\
Hangzhou, China\\
{\tt\small \{yuanyuanjia,tangshunpu,qianqianyang20\}@zju.edu.cn}}

\begin{document}
\maketitle
\vspace{-3em}
\makeatletter
\@ifpackageloaded{lineno}{\nolinenumbers}{}
\makeatother

\renewcommand{\thesection}{\Alph{section}}
\fi

\ifdefined\SUPPLEMENTINMAIN
  \newcommand{\suppwidebegin}{\begin{table*}[!t]\centering}
  \newcommand{\suppwidecaption}[1]{\caption{#1}}
  \newcommand{\suppwideend}{\end{table*}}
\else
  \newcommand{\suppwidebegin}{\begin{table*}[!t]\centering}
  \newcommand{\suppwidecaption}[1]{\caption{#1}}
  \newcommand{\suppwideend}{\end{table*}}
\fi

\ifdefined\SUPPLEMENTINMAIN
\begin{strip}
  \centering
  {\LARGE\bfseries Supplementary Material\par}
  \vspace{0.8em}
\end{strip}
\fi

\section{Additional Implementation Details}
\label{sec:supp_impl}

We implement $S^2$Prune on Qwen2.5-VL-7B-Instruct. 
For all experiments, each input image is resized to $672 \times 672$ using bicubic interpolation.
The vision encoder uses a patch size of 14 and a spatial merge factor of 2, producing a $48 \times 48$ feature grid before PatchMerger and a $24 \times 24$ grid after merging, corresponding to 576 visual tokens.

\subsection{Coarse Spatial Partition}
\label{sec:supp_partition}

For target visual-token budgets
$B \in \{32, 64, 128, 192\}$,
we partition the $24 \times 24$ visual-token grid into
$4 \times 4$, $5 \times 5$, $8 \times 8$, and $9 \times 9$ coarse grids, respectively.
We use $G$ to denote the total number of coarse regions; thus,
$G \in \{16,25,64,81\}$ for these four settings.
Region boundaries are obtained by rounding linearly spaced coordinates, which ensures complete coverage of all visual tokens without overlap.
When the token-grid size is not divisible by the coarse-grid side length, such as the $5 \times 5$ partition of a $24 \times 24$ grid, neighboring regions may differ in size by at most one token along each spatial dimension.

\subsection{Structural Complexity Computation}
\label{sec:supp_laplacian}

The full input image is first converted to a grayscale array $I$.
We apply one-pixel replicate padding at the outer image boundary and compute the four-neighbor discrete Laplacian response map over the entire image as
\begin{equation*}
    \begin{aligned}
    L(u,v)={}&I(u-1,v)+I(u+1,v)+I(u,v-1)\\
              &+I(u,v+1)-4I(u,v),
    \end{aligned}
\end{equation*}
where samples outside the image domain take the value of the nearest boundary pixel.

For each coarse region $R_g$, we map its token-grid boundaries proportionally to the image coordinates using rounded endpoints, obtaining the image-domain support $\Omega_g$.
We then crop the already computed response map as $L_g=L|_{\Omega_g}$ and compute the raw structural complexity as
\begin{equation*}
    c_g^{\mathrm{raw}}
    =
    \operatorname{Var}(L_g).
\end{equation*}

Thus, grayscale conversion, replicate padding, and Laplacian filtering are performed once on the full image, rather than independently within each regional crop.
Pixels along internal region boundaries therefore use their actual neighboring pixels from adjacent regions; replicate padding is used only at the outer image boundary.
For exact implementation, we distinguish the raw structural score from the allocation score.
Before regional token-budget allocation, the raw scores are independently min--max normalized within each image to obtain the allocation score
\begin{equation*}
    \tilde{c}_g
    =
    \frac{
        c_g^{\mathrm{raw}} - \min_j c_j^{\mathrm{raw}}
    }{
        \max\!\left(
            \max_j c_j^{\mathrm{raw}} - \min_j c_j^{\mathrm{raw}},
            \epsilon_{\mathrm{norm}}
        \right)
    },
\end{equation*}
where
\begin{equation*}
    \epsilon_{\mathrm{norm}}
    =
    \operatorname{finfo}(\mathrm{float32}).\mathrm{eps}
    \approx 1.1920929 \times 10^{-7}.
\end{equation*}
All reported experiments use $\tilde{c}_g$ for regional token-budget allocation.

\subsection{Regional Token-Budget Allocation}
\label{sec:supp_budget}

Each coarse region first receives one token to preserve broad spatial coverage.
The remaining $B-G$ tokens are distributed according to the normalized allocation scores $\tilde{c}_g$, subject to
\begin{equation*}
    1 \leq B_g \leq |R_g|,
    \qquad
    \sum_{g=1}^{G} B_g = B.
\end{equation*}

We enforce these constraints using an iterative largest-remainder allocation procedure.
At each iteration, only regions that have not reached their capacity are considered.
Let $\mathcal{A}$ denote the set of currently allocatable regions.
If
\begin{equation*}
    \sum_{g \in \mathcal{A}} \tilde{c}_g
    \leq
    \epsilon_{\mathrm{alloc}},
\end{equation*}
where
\begin{equation*}
    \epsilon_{\mathrm{alloc}}
    =
    \operatorname{finfo}(\mathrm{float64}).\mathrm{eps}
    \approx 2.2204460 \times 10^{-16},
\end{equation*}
we set the weights of all regions in $\mathcal{A}$ to one, which gives uniform allocation over the remaining regions.

Otherwise, the remaining budget is distributed proportionally to $\tilde{c}_g$.
If a region reaches its capacity $|R_g|$, any unassigned budget is redistributed among the remaining unsaturated regions.
For deterministic tie breaking in the largest-remainder step, regions are ordered by fractional remainder, allocation score, and finally row-major region index.
The procedure continues until exactly $B$ visual tokens have been allocated.

\subsection{Local Cell Construction}
\label{sec:supp_cells}

After obtaining the regional budget $B_g$, each coarse region $R_g$ is recursively partitioned into exactly $B_g$ non-empty, non-overlapping rectangular cells.

Initially, the entire coarse region forms a single rectangular cell.
At each iteration, we select the splittable rectangle containing the largest number of visual tokens.
If multiple rectangles have the same area, the rectangle appearing earliest in the current cell list is selected.

For a selected rectangle of height $H$ and width $W$, we split along the row dimension if
\begin{equation*}
    H \geq W
    \quad \text{and} \quad
    H > 1.
\end{equation*}
Otherwise, the rectangle is split along the column dimension.
Thus, square cells are preferentially split along rows.

For the selected dimension with interval starting at $s$ and length $L$, the split position is
\begin{equation*}
    m
    =
    s + \left\lfloor \frac{L}{2} \right\rfloor .
\end{equation*}
When $L$ is odd, the first child therefore contains
$\lfloor L/2 \rfloor$ rows or columns, while the second contains
$\lceil L/2 \rceil$.

This recursive procedure is repeated until exactly $B_g$ cells are obtained.
The resulting cells are all non-empty and mutually disjoint, and together they completely cover the original coarse region.
Each visual token therefore belongs to exactly one local cell.

\subsection{ERC-Based Local Selection}
\label{sec:supp_erc}

Before pruning, all 576 visual tokens are processed by decoder Layer~0.
For visual token $i$, the Early Representation Change (ERC) score is computed as
\begin{equation*}
    s_i
    =
    \left\|
        h_i^{1} - h_i^{0}
    \right\|_2,
\end{equation*}
where $h_i^{0}$ and $h_i^{1}$ denote the hidden representations immediately before and after decoder Layer~0, respectively.

The $\ell_2$ norm is computed in FP32 for numerical stability.
Within each local cell, we retain only the visual token with the largest ERC score.
Therefore, regional structural complexity determines how many tokens are allocated to each region, while ERC only determines which token represents each local cell.

\subsection{Token Ordering and Pruning}
\label{sec:supp_pruning}

$S^2$Prune does not modify the relative ordering of visual tokens before they enter the decoder.
Inside the Qwen2.5-VL vision tower, patch groups are temporarily reordered into window order for window attention and subsequently restored through the corresponding inverse permutation after PatchMerger.
This behavior is part of the official Qwen2.5-VL forward pass and is not introduced by $S^2$Prune.
Consequently, the full baseline and $S^2$Prune receive exactly the same 576 decoder-visible visual tokens in the same sequence order before pruning.

After token selection, the retained visual-token indices are sorted according to their original decoder-visible sequence indices.
Therefore, the pruned visual sequence is an order-preserving subsequence of the original sequence.
The retained tokens are never reordered according to Laplacian scores, ERC scores, or spatial-region indices.

Pruning is performed only once, immediately after decoder Layer~0 and before Layer~1.
Unselected visual tokens are physically removed from the sequence rather than merely masked out.
The hidden states, attention mask, Layer-0 KV cache, and the corresponding original M-RoPE position IDs are pruned consistently.

Although the sequence tensor becomes shorter after pruning, each retained visual token preserves its original pre-pruning M-RoPE position IDs; the position IDs are not renumbered.
Text tokens and special tokens are never removed and retain their original relative ordering and positional semantics.

\subsection{Training-Free Setting and Evaluation Consistency}
\label{sec:supp_eval_consistency}

$S^2$Prune introduces no trainable parameters.
It does not use query attention, text relevance, global semantic Top-$K$ selection, or dataset-specific parameter tuning.

For all comparisons, we use the same model checkpoint, image preprocessing, prompts, generation configuration, sample ordering, and evaluation pipeline across all methods.

\begin{algorithm}[t]
\caption{S$^2$Prune}
\label{alg:s2prune}
\footnotesize
\begin{algorithmic}[1]
\Require Image $I$; visual tokens $\mathcal{V}=\{v_i\}_{i=1}^{N}$ on an
$H\times W$ grid; target budget $B$; coarse-grid side length $K$, with
$G=K^2$ total regions
\Ensure Retained visual tokens $\mathcal{S}$ with $|\mathcal{S}|=B$
\State $\{\mathcal{R}_g\}_{g=1}^{G}\gets\Call{CoarsePartition}{H,W,K}$
\Comment{Partition the visual-token grid}
\State $I_{\mathrm{gray}}\gets\Call{Grayscale}{I}$
\State $L\gets\Call{Laplacian4N}{I_{\mathrm{gray}}}$
\Comment{Replicate padding}
\For{$g=1,\ldots,G$}
\State $\Omega_g\gets\Call{MapToImage}{\mathcal{R}_g}$
\Comment{Map token region to the original image}
\State $c_g\gets\operatorname{Var}(L[\Omega_g])$
\Comment{Regional structural complexity}
\State $n_g\gets|\mathcal{R}_g|$
\EndFor
\State $\widetilde{\mathbf c}\gets
\Call{MinMaxNormalize}{\mathbf c}$
\State $\{B_g\}_{g=1}^{G}\gets
\Call{AllocateBudget}{\widetilde{\mathbf c},\mathbf n,B}$
\Comment{$1\leq B_g\leq n_g$, $\sum_g B_g=B$}
\State $\mathbf H^1\gets
\Call{DecoderLayer0}{\mathbf H^0}$
\Comment{Process the full visual sequence}
\For{$i=1,\ldots,N$}
\State $s_i\gets
\left\|\mathbf H^1_{v_i}-\mathbf H^0_{v_i}\right\|_2$
\Comment{ERC score}
\EndFor
\State $\mathcal{S}\gets\emptyset$
\For{$g=1,\ldots,G$}
\State $\mathcal{Q}_g\gets
\Call{RecursivePartition}{\mathcal{R}_g,B_g}$
\ForAll{$Q\in\mathcal{Q}_g$}
\State $i^\star\gets
\arg\max_{i\in Q} s_i$
\State $\mathcal{S}\gets\mathcal{S}\cup\{i^\star\}$
\Comment{Select one token per cell}
\EndFor
\EndFor
\State $\mathcal{S}\gets\Call{SortByOriginalIndex}{\mathcal{S}}$
\Comment{Preserve the original visual-token order}
\State Physically remove unselected visual tokens after Layer~0
\State Preserve the original M-RoPE IDs of all retained tokens
\State Continue inference from decoder Layer~1
\State \Return $\mathcal{S}$
\end{algorithmic}
\end{algorithm}

\begin{algorithm}[t]
\caption{Regional Budget Allocation}
\label{alg:budget}
\footnotesize
\begin{algorithmic}[1]
\Require Normalized structural scores
$\widetilde{\mathbf c}=\{\tilde c_g\}_{g=1}^{G}$;
region capacities $\mathbf n=\{n_g\}_{g=1}^{G}$;
target budget $B$ satisfying $G\leq B\leq\sum_{g=1}^{G}n_g$
\Ensure Integer regional budgets
$\mathbf B=\{B_g\}_{g=1}^{G}$

\State $B_g\gets1,\quad \forall g$
\Comment{Assign one token to every region}
\State $R\gets B-G$
\Comment{Remaining token budget}

\While{$R>0$}
\State $\mathcal{A}\gets\{g:B_g<n_g\}$
\Comment{Regions with available capacity}
\If{$\sum_{g\in\mathcal{A}}\tilde c_g
    \leq\epsilon_{\mathrm{alloc}}$}
\State $w_g\gets1,\quad \forall g\in\mathcal{A}$
\Comment{Fall back to uniform allocation}
\Else
\State $w_g\gets\tilde c_g,\quad \forall g\in\mathcal{A}$
\EndIf

\State $q_g\gets
R\,w_g/\sum_{j\in\mathcal{A}}w_j,\quad \forall g\in\mathcal{A}$
\State $a_g\gets
\min\!\left(\lfloor q_g\rfloor,n_g-B_g\right),
\quad \forall g\in\mathcal{A}$
\State $B_g\gets B_g+a_g,\quad \forall g\in\mathcal{A}$
\State $R\gets B-\sum_g B_g$

\If{$R>0$}
\State Rank unsaturated regions by fractional remainder,
structural score, and region index
\ForAll{$g$ in ranked order}
\If{$R>0$ \textbf{and} $B_g<n_g$}
\State $B_g\gets B_g+1$
\State $R\gets R-1$
\EndIf
\EndFor
\EndIf
\EndWhile

\State \Return $\mathbf B$
\end{algorithmic}
\end{algorithm}

\begin{algorithm}[t]
\caption{Recursive Cell Partition}
\label{alg:cells}
\footnotesize
\begin{algorithmic}[1]
\Require Coarse region $R$; regional budget $B_g$ satisfying
$1\leq B_g\leq|R|$
\Ensure $B_g$ non-overlapping local cells $\mathcal{Q}$

\State $\mathcal{Q}\gets[R]$

\While{$|\mathcal{Q}|<B_g$}
\State $Q^\star\gets$ largest splittable rectangle in $\mathcal{Q}$
\Comment{Ties use the earliest cell}
\State Let $H$ and $W$ be the height and width of $Q^\star$

\If{$H\geq W$ \textbf{and} $H>1$}
\State Split $Q^\star$ along rows at $\lfloor H/2\rfloor$
\Else
\State Split $Q^\star$ along columns at $\lfloor W/2\rfloor$
\EndIf

\State Replace $Q^\star$ by the two resulting cells
\EndWhile

\State \Return $\mathcal{Q}$
\end{algorithmic}
\end{algorithm}

\section{Additional Spatial-Bias Results}
\label{sec:supp-spatial-bias}

On the $24\times24$ visual-token grid, the \emph{edge} region is the union of
the outermost two rows and the outermost two columns. It contains
$24^2-20^2=176$ positions, or approximately $30.6\%$ of the grid. The
\emph{center} region is the central half of the grid along both spatial axes,
which gives a $12\times12$ region. For a selected-token set $\mathcal{S}$, the
reported edge and center allocation ratios are the fractions of $\mathcal{S}$
whose grid positions fall in these regions. Beyond the Layer~3 result shown in
the main paper, delaying FEATHER's no-RoPE criterion to Layer~8 assigns
$67.6\%$ of retained tokens to the edge and $8.6\%$ to the
center. Thus, removing RoPE and pruning later does not eliminate the boundary
preference.

\begin{figure}[t]
  \centering
  \includegraphics[width=\columnwidth]{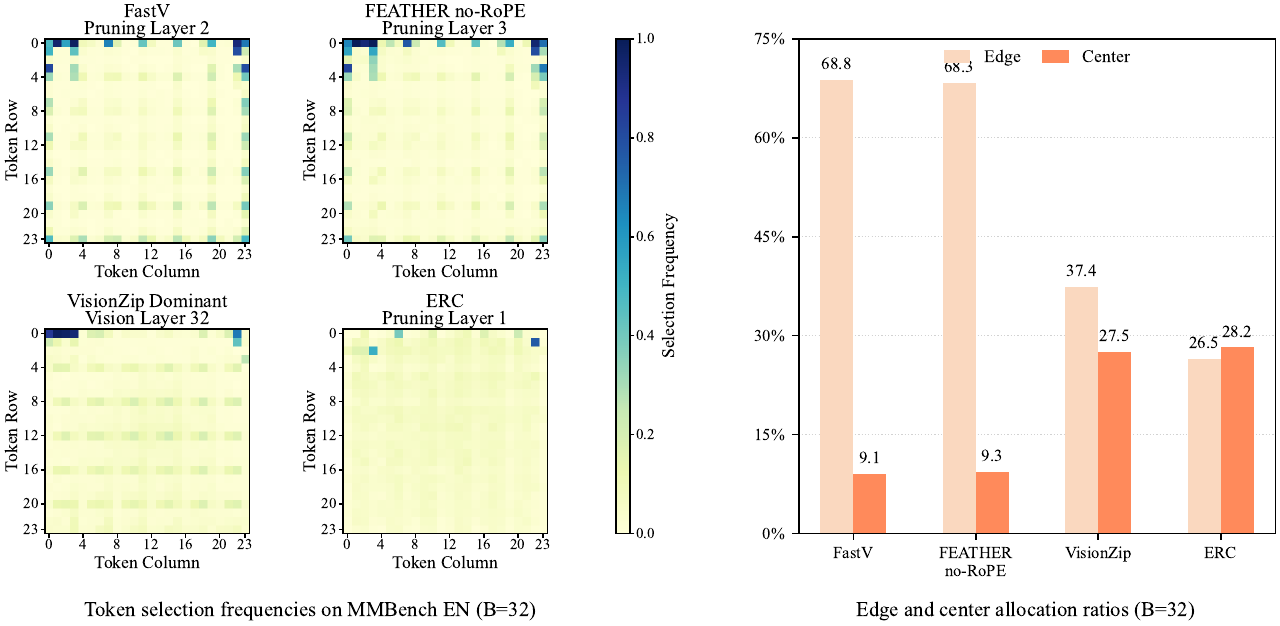}
  \caption{Spatial-bias measurements at $B=32$, extending Figures~2(b) and 2(c)
  of the main paper. The panel layout, 1,000 \datasetMMBEN{} samples, score
  definitions, and pruning criteria are unchanged from the main-paper analysis;
  only the retained-token budget differs.}
  \label{fig:supp-spatial-bias-b32}
\end{figure}

\begin{figure}[t]
  \centering
  \includegraphics[width=\columnwidth]{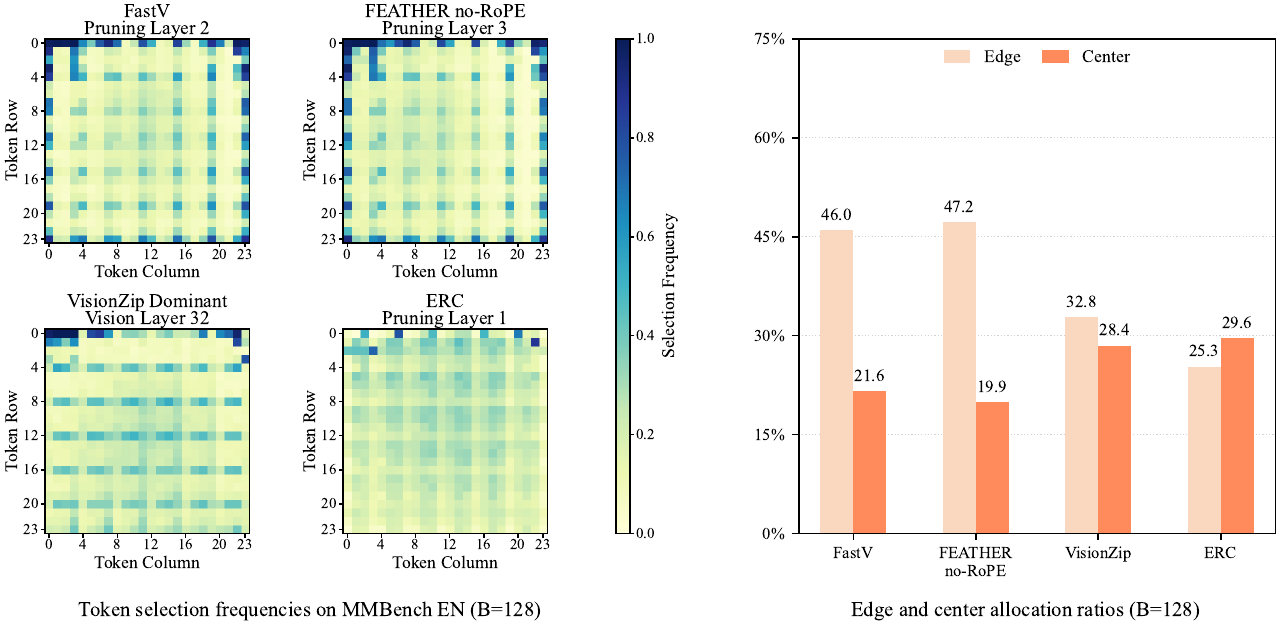}
  \caption{Spatial-bias measurements at $B=128$, extending Figures~2(b) and 2(c)
  of the main paper. The panel layout, 1,000 \datasetMMBEN{} samples, score
  definitions, and pruning criteria are unchanged from the main-paper analysis;
  only the retained-token budget differs.}
  \label{fig:supp-spatial-bias-b128}
\end{figure}

\paragraph{Stability across token budgets.}
\Cref{fig:supp-spatial-bias-b32,fig:supp-spatial-bias-b128} repeat the
main-paper measurements at $B=64$ for $B=32$ and $B=128$, using the same
1,000 \datasetMMBEN{} development samples and raw score definitions; only
Global Top-K changes. \Cref{tab:supp-spatial-correlation} shows Pearson and
Spearman correlations above $0.946$ for adjacent budgets. Even the
$32$-vs.-$128$ comparisons remain at or above $0.819$ and $0.889$, respectively,
showing stable positional patterns despite shifts in edge/center mass.

\begin{table}[H]
  \centering
  \caption{Cross-budget correlations between selection-frequency maps. Bold
  and underlined entries mark the best and second-best results within each
  budget pair, respectively.}
  \label{tab:supp-spatial-correlation}
  \scriptsize
  \setlength{\tabcolsep}{3.0pt}
  \renewcommand{\arraystretch}{0.96}
  \resizebox{0.92\columnwidth}{!}{%
  \begin{tabular}{lccc}
    \toprule
    Criterion & Budgets & Pearson $r$ & Spearman $\rho$ \\
    \midrule
    FastV & $32$ vs.\ $64$ & \textbf{0.9604} & 0.9478 \\
    FastV & $64$ vs.\ $128$ & \textbf{0.9624} & 0.9634 \\
    FastV & $32$ vs.\ $128$ & \textbf{0.8739} & 0.8899 \\
    \midrule
    FEATHER no-RoPE & $32$ vs.\ $64$ & 0.9480 & 0.9503 \\
    FEATHER no-RoPE & $64$ vs.\ $128$ & \underline{0.9596} & 0.9650 \\
    FEATHER no-RoPE & $32$ vs.\ $128$ & \underline{0.8415} & 0.8974 \\
    \midrule
    VisionZip & $32$ vs.\ $64$ & \underline{0.9504} & \textbf{0.9725} \\
    VisionZip & $64$ vs.\ $128$ & 0.9522 & \textbf{0.9797} \\
    VisionZip & $32$ vs.\ $128$ & 0.8269 & \textbf{0.9442} \\
    \midrule
    ERC & $32$ vs.\ $64$ & 0.9489 & \underline{0.9630} \\
    ERC & $64$ vs.\ $128$ & 0.9462 & \underline{0.9726} \\
    ERC & $32$ vs.\ $128$ & 0.8190 & \underline{0.9217} \\
    \bottomrule
  \end{tabular}%
  }
\end{table}

\section{Additional Qwen2.5-VL-7B Results}
\label{sec:supp-qwen-results}

\Cref{tab:supp-b192} reports the available $B=192$ evaluations.
POPE is reported with F1, and MMVet uses the local reference-match evaluation.

\suppwidebegin
  \suppwidecaption{Available Qwen2.5-VL-7B results at $B=192$. POPE is
  reported with F1, and MMVet uses the local reference-match evaluation.
  Acc./Rel. aggregate all ten benchmarks. Bold and
  underlined entries mark the best and second-best reported results per
  column, respectively; ties share the same style, and rankings use the
  unrounded scores.}
  \label{tab:supp-b192}
  \footnotesize
  \setlength{\tabcolsep}{2.3pt}
  \renewcommand{\arraystretch}{1.05}
  \resizebox{\textwidth}{!}{%
  \begin{tabular}{l|cccccccccc|cc}
    \toprule
    Method & GQA & \datasetSQA & \datasetVQAText & VizWiz & MMMU & POPE F1 & MME & \datasetMMBEN & \datasetMMBCN & MMVet & Acc. & Rel. \\
    \midrule
    FastV & 54.1 & 85.0 & \textbf{72.1} & \textbf{35.8} & 50.0 & 81.6 & \underline{2200} & 86.3 & 85.3 & 28.0 & 65.7 & 94.9\% \\
    DART & 55.3 & 83.3 & 60.5 & 32.7 & \underline{50.3} & 80.9 & 2165 & 86.0 & 85.4 & 25.2 & 63.7 & 91.6\% \\
    VisionZip & 54.7 & 84.7 & 62.1 & 34.0 & 49.3 & 80.4 & 2154 & \textbf{86.7} & 86.1 & 27.5 & 64.2 & 92.8\% \\
    GPrune & 54.0 & 84.4 & 59.2 & 30.9 & 48.6 & 81.4 & 2020 & 84.6 & 84.1 & 24.8 & 62.4 & 89.5\% \\
    VisPruner & 54.8 & \textbf{85.5} & 69.1 & 34.8 & 49.7 & \underline{82.2} & 2197 & 86.1 & 85.5 & 28.4 & 65.5 & 94.6\% \\
    GridPrune & 55.3 & 84.7 & 67.4 & \underline{35.2} & \textbf{51.0} & 82.1 & 2194 & 85.9 & \textbf{86.4} & 25.2 & 65.2 & 93.8\% \\
    HoloV & 55.2 & \textbf{85.5} & 69.4 & 34.8 & 47.7 & \textbf{82.7} & 2192 & \underline{86.4} & 86.1 & 29.4 & 65.5 & 94.8\% \\
    \rowcolor{resultours}
    \textbf{\methodname ($9\times9$)} & \underline{56.3} & 85.2 & \underline{69.6} & 34.7 & 48.8 & 81.9 & \textbf{2233} & \underline{86.4} & \underline{86.3} & \underline{30.3} & \textbf{65.9} & \textbf{95.5\%} \\
    \rowcolor{resultours}
    \textbf{\methodname ($10\times10$)} & \textbf{56.5} & \underline{85.3} & 69.4 & 34.1 & 48.9 & 81.3 & 2197 & \textbf{86.7} & 86.2 & \textbf{30.7} & \underline{65.8} & \underline{95.3\%} \\
    \bottomrule
  \end{tabular}%
  }
\suppwideend
\ifdefined\SUPPLEMENTINMAIN
  \FloatBarrier
\fi

Following the main paper, Acc. is the
arithmetic mean of the percentage-format benchmark scores (with MME divided
by 28), while Rel. is the mean per-benchmark retention relative to the
full-token score. MME is displayed as the nearest integer, while Acc./Rel. use
the underlying unrounded MME values. The table contains complete results for all ten
benchmarks, so its Acc./Rel. columns use the same ten-benchmark aggregate.

\paragraph{Analysis.}
At $B=192$, the $9\times9$ configuration achieves the best aggregate performance, with $65.9$ Acc. and $95.5\%$ Rel., compared with $65.7$ Acc. and $94.9\%$ Rel. for the strongest baseline, FastV.
The $10\times10$ configuration ranks second on both aggregate metrics, showing that the overall performance is stable under a moderate change in grid size.
Across individual benchmarks, the $9\times9$ configuration performs best on MME, while the $10\times10$ configuration performs best on GQA, \datasetMMBEN, and MMVet; its \datasetMMBEN{} result ties VisionZip.
The gains are not uniform across all datasets: FastV remains strongest on \datasetVQAText{} and VizWiz, GridPrune on MMMU and \datasetMMBCN, HoloV on POPE, and HoloV and VisPruner tie on \datasetSQA.
Overall, the $B=192$ results show that \methodname achieves the strongest aggregate performance, although the leading method varies across individual datasets.

\section{Grid-Size Sensitivity}
\label{sec:supp-grid-size}

\Cref{tab:supp-grid-size} compares the coarse-grid configurations evaluated
at token budgets $B=64$ and $B=32$. Acc./Rel. aggregate all four reported
benchmarks using the same definitions as above.

\begin{table}[H]
  \centering
  \caption{Grid-size sensitivity of \methodname on Qwen2.5-VL-7B. Acc./Rel.
  aggregate all four benchmarks. Bold and underlined entries mark the best
  and second-best results within each token budget, respectively; rankings
  use the unrounded scores.}
  \label{tab:supp-grid-size}
  \small
  \setlength{\tabcolsep}{3.0pt}
  \renewcommand{\arraystretch}{1.05}
  \resizebox{0.92\columnwidth}{!}{%
  \begin{tabular}{lcccc|cc}
    \toprule
    Budget / Grid & \datasetMMBEN & \datasetVQAText & \datasetSQA & POPE & Acc. & Rel. \\
    \midrule
    $B=64$ / $5\times5$ & 84.6 & \textbf{55.2} & 81.0 & \textbf{74.8} & 73.9 & 86.3\% \\
    $B=64$ / $4\times4$ & \textbf{85.4} & \underline{54.6} & 81.4 & 74.5 & \textbf{74.0} & \textbf{86.4\%} \\
    $B=64$ / $6\times6$ & 84.9 & 53.8 & \textbf{82.4} & \underline{74.8} & \underline{73.9} & \underline{86.3\%} \\
    $B=64$ / $8\times8$ & \underline{85.3} & 52.9 & \underline{81.5} & 74.0 & 73.4 & 85.7\% \\
    \midrule
    $B=32$ / $4\times4$ & \underline{82.6} & \underline{44.9} & \underline{78.4} & \underline{66.9} & \underline{68.2} & \underline{79.5\%} \\
    $B=32$ / $3\times3$ & 82.0 & 44.4 & 77.6 & 66.7 & 67.7 & 78.9\% \\
    $B=32$ / $5\times5$ & \textbf{82.8} & \textbf{44.9} & \textbf{79.0} & \textbf{67.1} & \textbf{68.5} & \textbf{79.8\%} \\
    \bottomrule
  \end{tabular}%
  }
\end{table}

\paragraph{Analysis.}
The results show that performance is only mildly affected by reasonable changes in the coarse-grid size.
At $B=64$, the $4\times4$, $5\times5$, and $6\times6$ grids differ by at most $0.1$ aggregate Acc.; at $B=32$, the $3\times3$, $4\times4$, and $5\times5$ grids remain within $0.8$ Acc.
We therefore treat the grid size as an empirical, budget-dependent setting rather than a dataset-specific hyperparameter, and use one fixed configuration across all benchmarks for each budget.
Specifically, we use $4\times4$, $5\times5$, $8\times8$, and $9\times9$ grids for $B=32$, $64$, $128$, and $192$, respectively.

\begin{figure*}[!t]
  \centering
  \includegraphics[page=1,width=\textwidth]{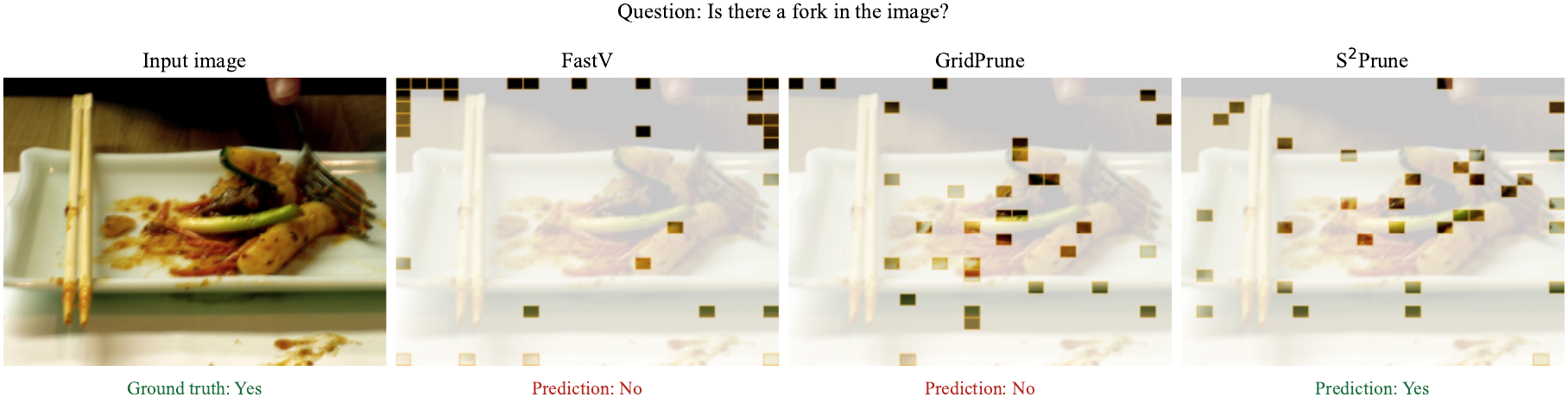}\par\vspace{0.35em}
  \includegraphics[page=2,width=\textwidth]{figure/qualitative_examples_multipage.pdf}
  \caption{At $B=32/576$, \methodname answers both POPE queries correctly,
  whereas FastV and GridPrune answer incorrectly. The queries concern a fork
  and couch; green and red text indicate correct and incorrect predictions,
  respectively.}
  \label{fig:supp-qualitative-1}
\end{figure*}

\section{Qualitative Results}
\label{sec:supp-qualitative}

\begin{figure*}[!t]
  \centering
  \includegraphics[page=3,width=\textwidth]{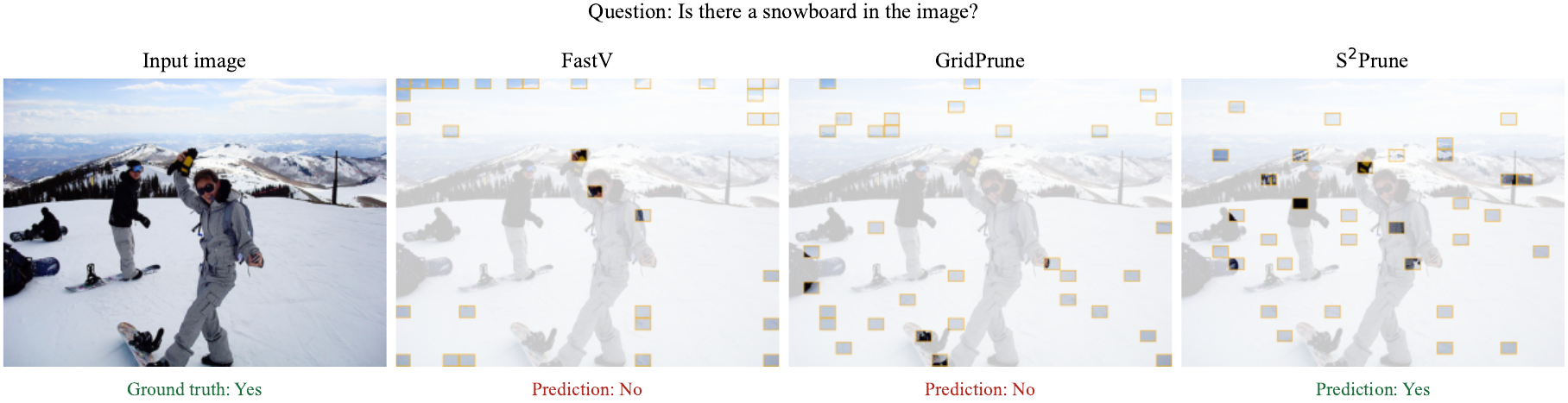}\par\vspace{0.35em}
  \includegraphics[page=4,width=\textwidth]{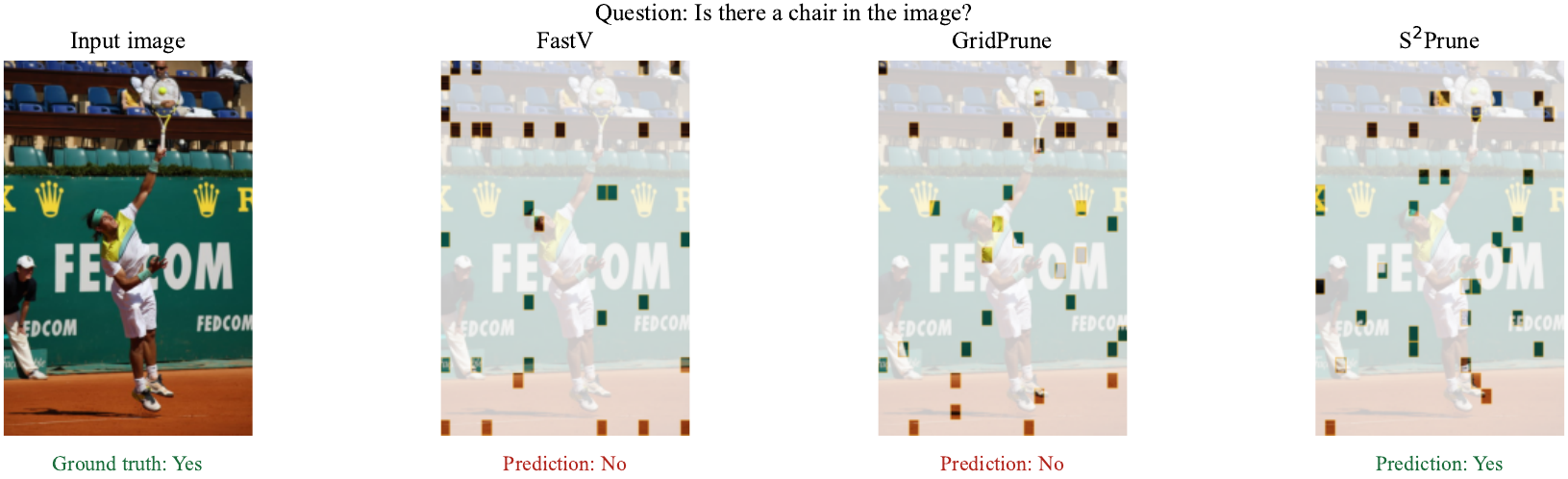}\par\vspace{0.35em}
  \includegraphics[page=5,width=\textwidth]{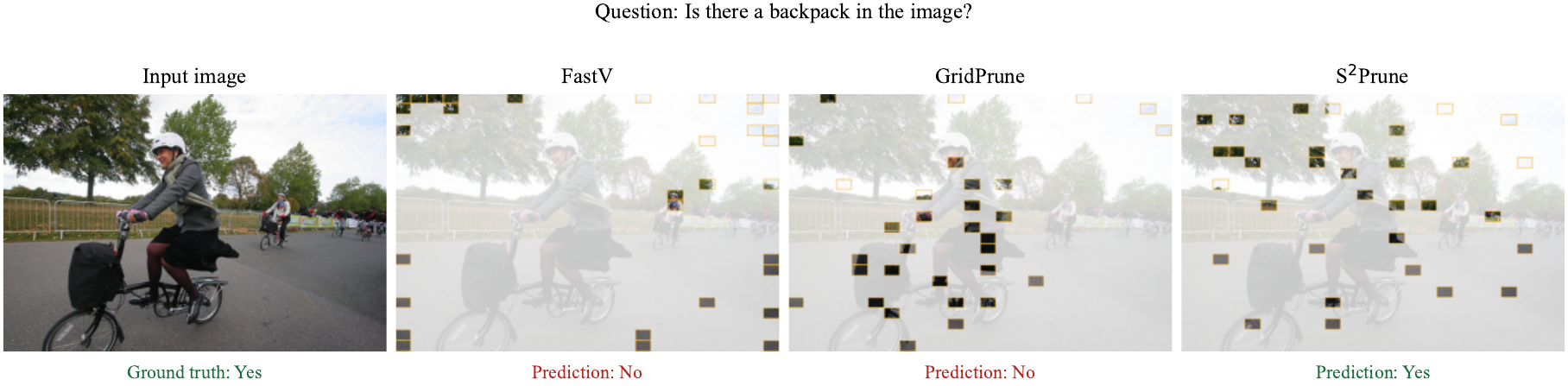}\par\vspace{0.35em}
  \includegraphics[page=6,width=\textwidth]{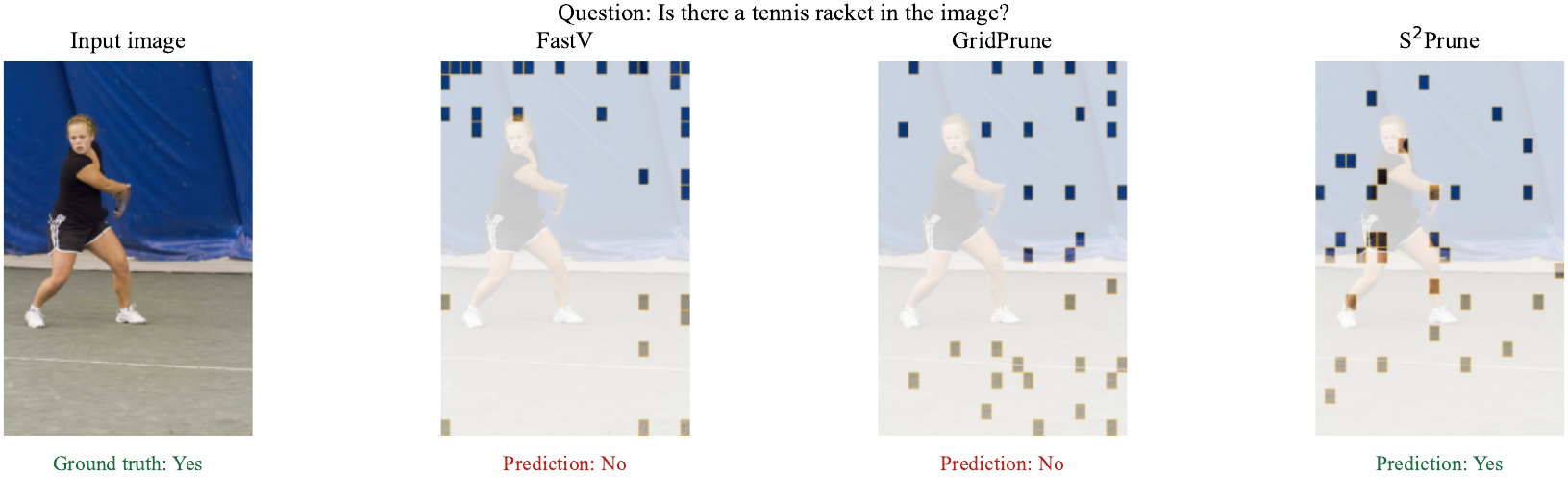}
  \caption{At $B=32/576$, \methodname answers these four additional POPE
  queries correctly, whereas FastV and GridPrune answer incorrectly. The
  queries concern a snowboard, chair, backpack, and tennis racket; the layout
  and color conventions match \cref{fig:supp-qualitative-1}.}
  \label{fig:supp-qualitative-2}
\end{figure*}

\begin{figure*}[!t]
  \centering
  \includegraphics[page=1,width=\textwidth]{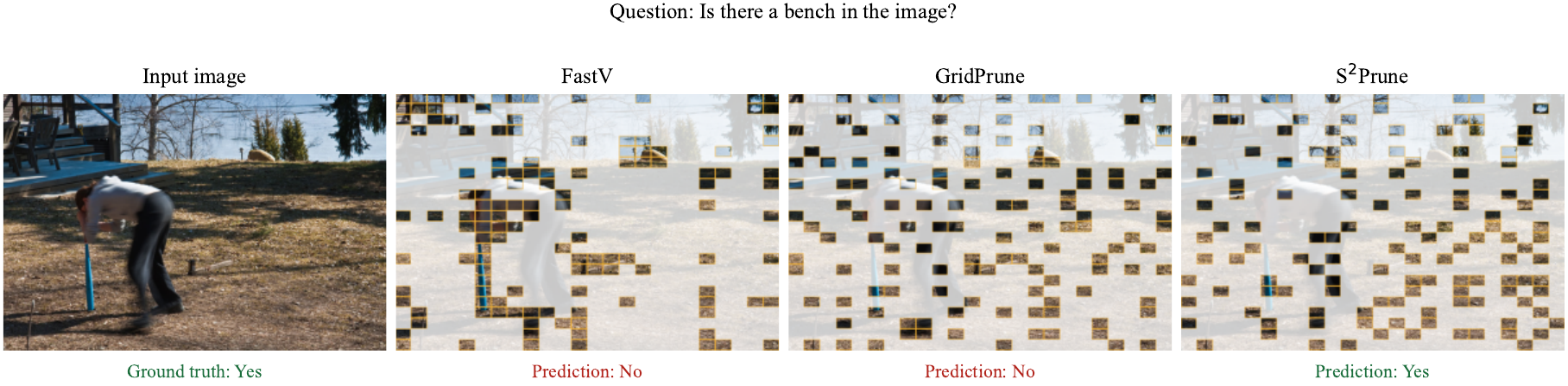}\par\vspace{0.35em}
  \includegraphics[page=2,width=\textwidth]{figure/qualitative_examples_multipage128.pdf}\par\vspace{0.35em}
  \includegraphics[page=3,width=\textwidth]{figure/qualitative_examples_multipage128.pdf}
  \caption{At $B=128/576$, \methodname answers all three POPE queries correctly,
  whereas FastV and GridPrune answer incorrectly. The queries concern a bench,
  vase, and car; green and red text indicate correct and incorrect predictions,
  respectively.}
  \label{fig:supp-qualitative-b128-1}
\end{figure*}

\begin{figure*}[!t]
  \centering
  \includegraphics[page=4,width=\textwidth]{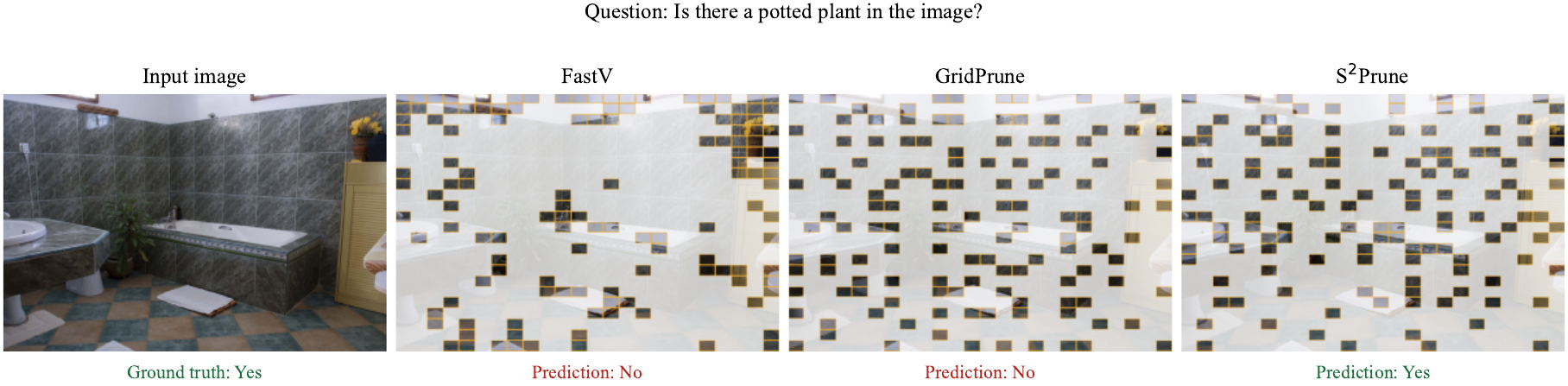}\par\vspace{0.35em}
  \includegraphics[page=5,width=\textwidth]{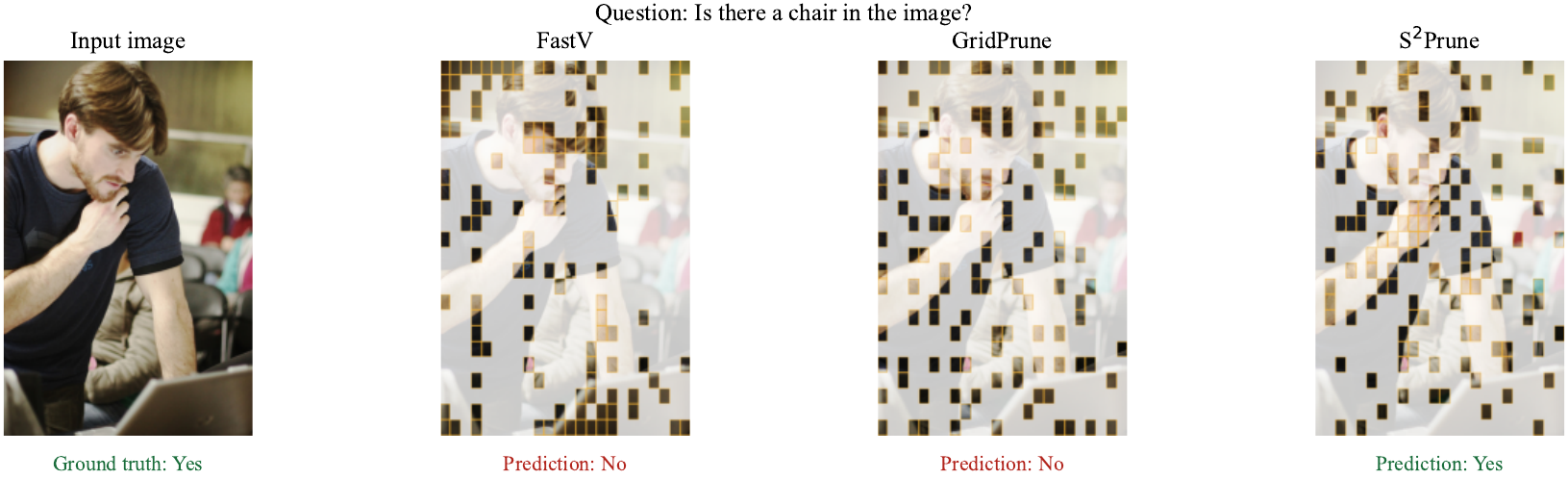}\par\vspace{0.35em}
  \includegraphics[page=6,width=\textwidth]{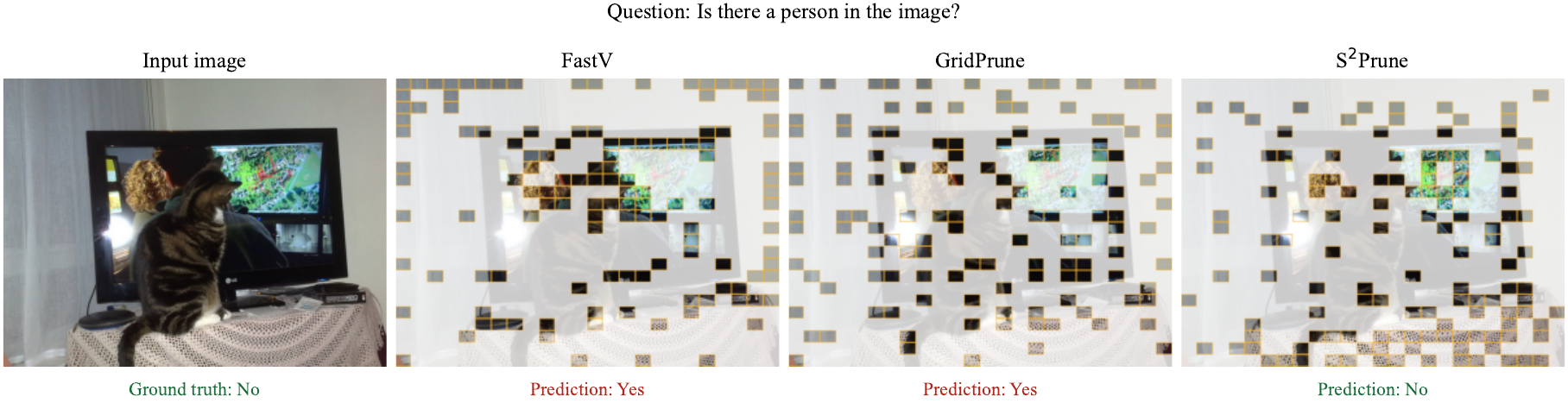}
  \caption{At $B=128/576$, \methodname answers these three additional POPE
  queries correctly, whereas FastV and GridPrune answer incorrectly. The
  queries concern a potted plant, chair, and person; the layout and color
  conventions match \cref{fig:supp-qualitative-b128-1}.}
  \label{fig:supp-qualitative-b128-2}
\end{figure*}

\Cref{fig:supp-qualitative-1,fig:supp-qualitative-2} show POPE examples for
Qwen2.5-VL-7B-Instruct with $B=32$ retained visual tokens out of 576, while
\cref{fig:supp-qualitative-b128-1,fig:supp-qualitative-b128-2} show additional
examples with $B=128$ retained tokens. In both settings, we compare the visual
tokens retained by FastV, GridPrune, and \methodname.

\paragraph{Setup.}
All six $B=32/576$ cases are positive POPE samples whose ground-truth answer is
``Yes.'' The $B=128/576$ cases contain five positive samples and one negative
sample, corresponding to the person query in
\cref{fig:supp-qualitative-b128-2}. In all shown cases, FastV and GridPrune
produce an incorrect answer, whereas \methodname produces the correct answer.
The overlays visualize the retained visual-token positions and therefore make
it possible to compare not only the final answer but also the spatial evidence
preserved by each method.

\paragraph{Observation.}
The queried evidence spans different scales and shapes. The fork and tennis
racket are thin objects; the snowboard is partly surrounded by visually
similar snow; and the couch, chair, and backpack appear amid people and other
scene content. FastV often spends tokens near image boundaries, while
GridPrune provides broader spatial coverage but can still leave limited local
evidence on the queried object. In these examples, \methodname retains a more
spatially distributed set of tokens together with local evidence around the
target object. This behavior is obtained without query-conditioned scoring:
regional image structure determines the allocation, and ERC selects the local
representative within each cell. These selected cases illustrate the spatial
behavior of the methods and complement, rather than replace, the aggregate
POPE evaluation.

\FloatBarrier

\ifdefined\SUPPLEMENTINMAIN
  \let\supplementend\relax
\else
  \def\supplementend{\end{document}}
\fi
\supplementend

\end{document}